\documentclass{article}

\usepackage{microtype}
\usepackage{graphicx}
\usepackage{placeins}
\usepackage{float}
\usepackage{colortbl}
\usepackage{subcaption}
\usepackage{booktabs} % for professional tables
\usepackage{multirow} % for multi-row cells in tables
\usepackage{color, soul}
\usepackage{xcolor}

\usepackage{hyperref}

\usepackage[preprint]{icml2026}

\usepackage{amsmath}
\usepackage{amssymb}
\usepackage{mathtools}
\usepackage{amsthm}
\usepackage{tikz}

\usepackage[capitalize,noabbrev]{cleveref}

\usepackage{placeins}

\theoremstyle{plain}

\theoremstyle{definition}

\theoremstyle{remark}

\usepackage[textsize=tiny]{todonotes}

\icmltitlerunning{ScoRe-Flow: Complete Distributional Control for Flow Matching}

\begin{document}
\raggedbottom

\twocolumn[
  \icmltitle{ScoRe-Flow: Complete Distributional Control via Score-Based Reinforcement Learning for Flow Matching}

  % It is OKAY to include author information, even for blind submissions: the
  % style file will automatically remove it for you unless you've provided
  % the [accepted] option to the icml2026 package.

  % List of affiliations: The first argument should be a (short) identifier you
  % will use later to specify author affiliations Academic affiliations
  % should list Department, University, City, Region, Country Industry
  % affiliations should list Company, City, Region, Country

  % You can specify symbols, otherwise they are numbered in order. Ideally, you
  % should not use this facility. Affiliations will be numbered in order of
  % appearance and this is the preferred way.
  \icmlsetsymbol{equal}{*}

  \begin{icmlauthorlist}
    \icmlauthor{Xiaotian Qiu}{equal,zju,sii}
    \icmlauthor{Lukai Chen}{equal,sjtu}
    \icmlauthor{Jinhao Li}{sjtu,sii}
    \icmlauthor{Qi Sun}{zju}
    \icmlauthor{Cheng Zhuo}{zju}
    \icmlauthor{Guohao Dai}{sjtu,inf,sii}
  \end{icmlauthorlist}

  \icmlaffiliation{zju}{Zhejiang University, Hangzhou, China}
  \icmlaffiliation{sjtu}{Shanghai Jiao Tong University, Shanghai, China}
  \icmlaffiliation{sii}{Shanghai Innovation Institute}
  \icmlaffiliation{inf}{Infinigence-AI}

  \icmlcorrespondingauthor{Guohao Dai}{daiguohao@sjtu.edu.cn}

  % You may provide any keywords that you find helpful for describing your
  % paper; these are used to populate the "keywords" metadata in the PDF but
  % will not be shown in the document
  \icmlkeywords{Machine Learning, ICML}

  \vskip 0.3in
]

% this must go after the closing bracket ] following \twocolumn[ ...

% This command actually creates the footnote in the first column listing the
% affiliations and the copyright notice. The command takes one argument, which
% is text to display at the start of the footnote. The \icmlEqualContribution
% command is standard text for equal contribution. Remove it (just {}) if you
% do not need this facility.

% Use ONE of the following lines. DO NOT remove the command.
% If you have no special notice, KEEP empty braces:
\printAffiliationsAndNotice{\icmlEqualContribution Work done during an internship at Shanghai Jiao Tong University.}

\begin{abstract}
Flow Matching (FM) policies have emerged as an efficient backbone for robotic control, offering fast and expressive action generation that underpins recent large-scale embodied AI systems. However, FM policies trained via imitation learning inherit the limitations of demonstration data; surpassing suboptimal behaviors requires reinforcement learning (RL) fine-tuning. Recent methods convert deterministic flows into stochastic differential equations (SDEs) with learnable noise injection, enabling exploration and tractable likelihoods, but such noise-only control can compromise training efficiency when demonstrations already provide strong priors. We observe that modulating the drift via the score function, i.e., the gradient of log-density, steers exploration toward high-probability regions, improving stability. The score admits a closed-form expression from the velocity field, requiring no auxiliary networks. Based on this, we propose \textbf{ScoRe-Flow}, a score-based RL fine-tuning method that combines drift modulation with learned variance prediction to achieve decoupled control over the mean and variance of stochastic transitions. Experiments demonstrate that ScoRe-Flow achieves \textbf{2.4$\times$ faster convergence} than flow-based SOTA on D4RL locomotion tasks and \textbf{up to 5.4\% higher success rates} on Robomimic and Franka Kitchen manipulation tasks.
\end{abstract}

\section{Introduction}

Generative policies have reshaped robotic control by moving beyond unimodal regression toward modeling rich multimodal action distributions that better reflect the variability of human demonstrations \citep{florence2022implicit, shafiullah2022behavior, pearce2023imitating, braunriemannian, zhang2024affordance}. This capability is essential for complex manipulation tasks where multiple valid action sequences can achieve the same goal, and the policy must capture this inherent ambiguity rather than averaging diverse behaviors. Among recent approaches, diffusion-based policies \citep{chi2023diffusion, reuss2023goal} achieve strong performance through stochastic sampling but often require many iterative denoising steps, limiting their suitability for real-time control and long-horizon decision making.

\begin{figure}[t]
    \centering
    \includegraphics[width=\columnwidth]{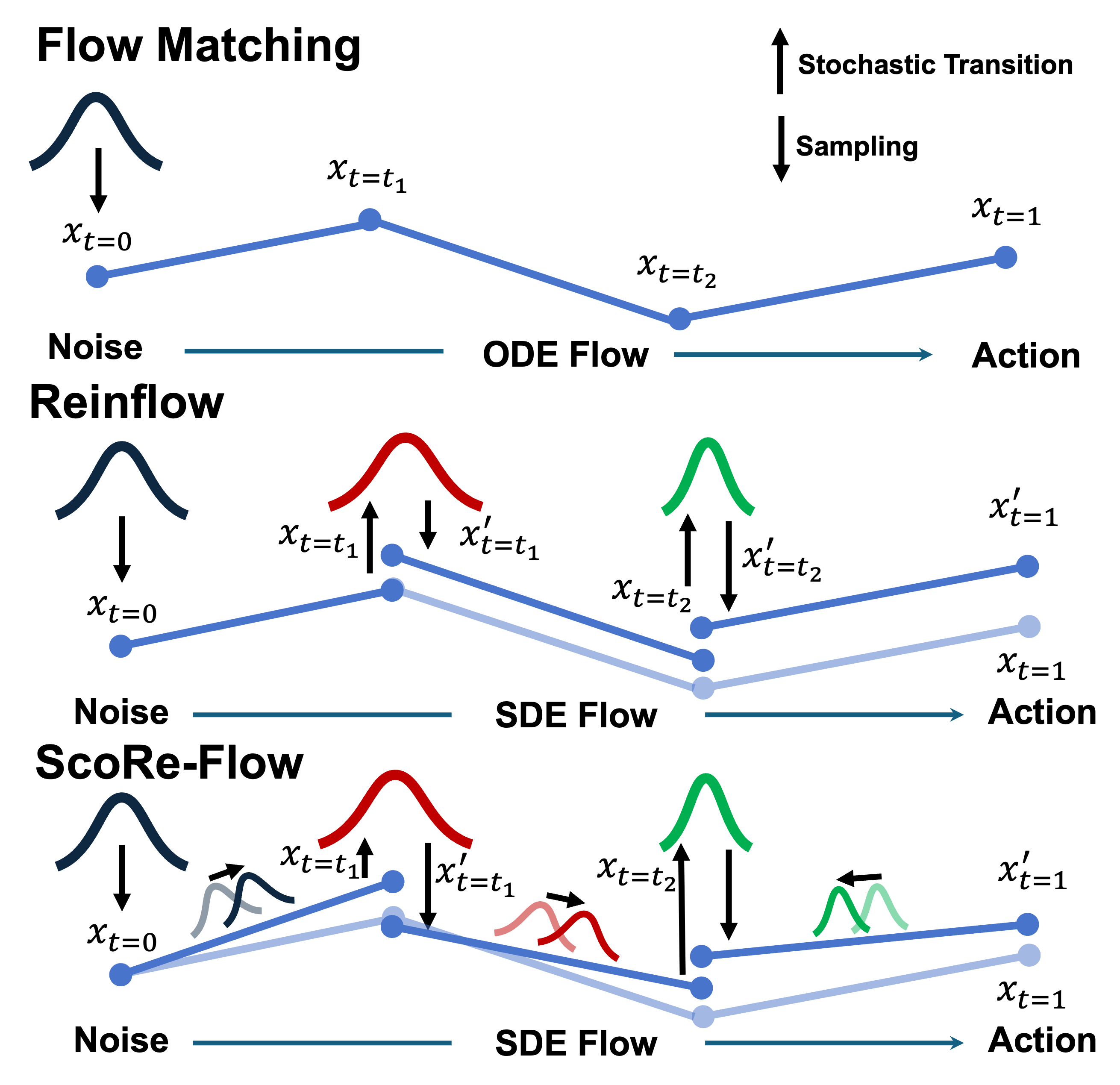}
    \caption{Comparison of flow policy sampling strategies. \textbf{Top}: Deterministic FM follows a fixed ODE trajectory with no exploration. \textbf{Middle}: Noise-only control (e.g., ReinFlow) injects learnable noise for exploration, but only perturbs the position without modulating dynamics. \textbf{Bottom}: ScoRe-Flow combines score-based drift modulation with learned variance prediction, achieving decoupled mean-variance control.}
    \label{fig:overview}
\end{figure}

Flow Matching (FM) provides an ODE-based formulation for action generation, transporting noise to actions along structured trajectories \citep{lipman2022flow, liu2022flow, albergo2023stochastic}. Compared to diffusion models, FM enables significantly faster sampling, typically requiring 4 to 10$\times$ fewer integration steps, while maintaining comparable expressiveness. This formulation has recently been adopted in several large-scale robotic action experts and vision-language-action models \citep{black2024pi_0, amin2025pi, chen2025pi_}.

However, most FM policies rely solely on imitation learning (IL) \citep{albergo2025stochastic}, inherently suffering from covariate shift and the performance ceiling of suboptimal demonstrations \citep{ross2011reduction}. 
Although RL fine-tuning is essential to transcend these limitations \citep{rajeswaran2017learning}, a fundamental challenge arises from the deterministic ODE dynamics of FM: transitions collapse to Dirac measures \citep{song2020score}, rendering action likelihoods ill-defined and precluding direct application of policy gradient methods \citep{zhang2025reinflow}.

To enable RL fine-tuning of FM policies, recent methods convert deterministic flows into stochastic differential equations (SDEs) with tractable likelihoods \citep{black2023training, fan2024dppo}. A representative approach is ReinFlow \citep{zhang2025reinflow}, which learns an auxiliary network to predict the magnitudes of injected noise at each step of integration. Other concurrent strategies focus on optimizing the input noise distribution to discover high-reward seeds \citep{yu2025smart} or modifying the integrator to mitigate sampling artifacts \citep{wang2025coefficients}. Although these methods enable exploration and stabilize fine-tuning, they primarily act through the \textit{noise} component. Concretely, under Euler-Maruyama discretization \citep{oksendal2003stochastic}, the update $\mathbf{a}_{k+1} = \mathbf{a}_k + \mathbf{v}\Delta t + \sigma\epsilon$ shows that injected noise perturbs the state increment but leaves the drift $\mathbf{v}$ unchanged. This motivates drift modulation as a complementary mechanism for providing directional guidance during sampling.

In this work, we introduce learnable drift modulation for RL fine-tuning of flow policies. Building on the known score-velocity duality for linear flow paths \citep{albergo2023stochastic, lipman2022flow}, we observe that the density gradient (score) can be computed in closed form from the velocity field, enabling score-based drift control without auxiliary networks. Based on this, we propose \textbf{ScoRe-Flow} (\textbf{Sco}re-based \textbf{Re}inforcement Learning for \textbf{Flow} Matching), combining score-weighted drift modulation with variance prediction to achieve decoupled control over the mean and variance of stochastic transitions. Empirically, this yields faster convergence on D4RL and improved performance on Robomimic and Franka Kitchen.

In summary, our contributions are threefold as follows:
\begin{itemize}
\item We observe that noise-based SDE conversion for FM fine-tuning only perturbs the \textbf{position} at each step without modulating the \textbf{dynamics} of the flow. The score function, which points toward high-probability regions, provides a principled mechanism to adjust the drift and steer transitions along geometrically favorable paths.

\item Leveraging the known score-velocity duality for linear flow paths \citep{albergo2023stochastic, lipman2022flow}, we propose \textbf{ScoRe-Flow}, which uses the \textbf{closed-form score} to enable drift modulation requiring \textbf{no auxiliary networks and negligible computational overhead}. Combined with variance prediction, this achieves \textbf{decoupled control} over the mean and variance of stochastic transitions — a capability absent in existing flow-based RL methods.

\item We demonstrate that ScoRe-Flow achieves \textbf{2.4$\times$ faster convergence} than the flow-based SOTA (ReinFlow) at matched denoising steps, and \textbf{up to 5.4\% higher success rates} on sparse-reward Robomimic and Franka Kitchen manipulation tasks. The additional wall-clock advantage over diffusion-based methods (21.9$\times$ over DPPO) is a structural benefit of flow-based sampling shared by all FM methods. Results are validated across state-based and vision-based settings.
\end{itemize}

\section{Related Work}

\paragraph{Generative policies for robotic control.}
Generative models have emerged as powerful backbones for robotic policy learning, offering the ability to capture multimodal action distributions that reflect the inherent variability in human demonstrations \citep{florence2022implicit, shafiullah2022behavior, pearce2023imitating}. Diffusion-based policies \citep{chi2023diffusion, janner2022planning, reuss2023goal} achieve strong performance through iterative denoising, but often require many sampling steps, limiting the applicability in real-time. Flow Matching (FM) \citep{lipman2022flow, liu2022flow, albergo2023stochastic} provides an ODE-based alternative by learning deterministic velocity fields that transport noise to actions along structured trajectories, enabling significantly faster sampling (often 4 to 10$\times$ fewer steps). Recent large-scale robotic systems have adopted FM as the action generation backbone \citep{black2024pi_0, amin2025pi, chen2025pi_}.

\paragraph{RL fine-tuning of diffusion and flow policies.}
Although imitation learning provides robust initialization, reinforcement learning (RL) fine-tuning is essential to exceed demonstration quality \citep{ball2023efficient, ibarz2021train}. 
However, integrating expressive generative policies with RL objectives presents significant challenges, and existing approaches can be broadly categorized by their optimization strategies.

\emph{Value-based and Energy-based approaches} leverage learned Q-functions or energy landscapes to guide action selection, though these methods typically prioritize offline settings or implicit policy formulations \citep{hansen2023idql, park2025flow, zhang2025energy}. 
Similarly, other works propose structural modifications to incorporate flow-based networks into actor-critic architectures \citep{zhong2025flowcritic, bai2025pacer}, tackling distributional matching via architectural redesigns which is different from our objective of enhancing sample efficiency in online fine-tuning through geometric guidance of the flow trajectory

% \emph{Value-based approaches} leverage Q-functions to guide action selection. IDQL \citep{hansen2023idql} treats diffusion models as implicit policies and uses Q-values for weighted sampling. FQL \citep{park2025flow} extends this to flow policies but lacks explicit exploration mechanisms for online adaptation. Value-guided methods \citep{dong2025value, zheng2023guided} use critic gradients to steer generation but may require differentiating through multi-step denoising.

\emph{Policy gradient approaches} directly optimize the generative policy using likelihood-based objectives. DPPO \citep{ren2024diffusion} applies PPO to diffusion policies by modeling denoising as a multi-step MDP. For flow policies, ReinFlow \citep{zhang2025reinflow} converts the deterministic ODE into an SDE by learning noise injection. Smart-GRPO \citep{yu2025smart} optimizes the input noise distribution to discover high-reward seeds, while Coefficients-Preserving Sampling \citep{wang2025coefficients} modifies the integrator to reduce sampling artifacts. 
However, these approaches largely focus on optimizing noise parameters or stabilizing numerical integration, rather than leveraging the underlying probability geometry. 
Additionally, unlike language-domain approaches like FlowRL \citep{zhu2025flowrl}, which prioritize diversity through distribution matching, our framework is tailored for robotic control, focusing on maximizing expected returns to ensure precise and reliable manipulation.
% \emph{Policy gradient approaches} directly optimize generative policies via likelihood-based objectives, employing strategies such as multi-step MDP formulations, SDE conversions, or input noise optimization \citep{ren2024diffusion, zhang2025reinflow, yu2025smart, wang2025coefficients}. 
% However, these methods predominantly focus on managing noise components or integrator mechanics, distinct from our approach of actively steering the flow dynamics via score-based drift modulation. 
% Furthermore, while distribution matching has been explored to encourage diversity in language reasoning \citep{zhu2025flowrl}, our framework targets the specific requirement of maximizing expected returns for precise robotic control.

% \emph{Policy gradient approaches} directly optimize the generative policy using likelihood-based objectives. DPPO \citep{ren2024diffusion} applies PPO to diffusion policies by modeling denoising as a multi-step MDP, though with high computational cost. For flow policies, ReinFlow \citep{zhang2025reinflow} converts the deterministic ODE into an SDE by learning noise injection, enabling tractable likelihood computation and exploration. Recent works explore alternative formulations including GRPO-based methods \citep{liu2025flow, li2025mixgrpo, he2025tempflow} and flow matching policy gradients \citep{mcallister2025flow}. Our method belongs to this category and builds upon the SDE conversion framework, introducing score-based drift modulation for improved training dynamics.

\paragraph{Score functions and guided generation.}
The score function, i.e., the gradient of log-density, plays a central role in diffusion models \citep{song2019generative, song2020score} and connects to various guidance techniques for conditional generation \citep{ho2022classifier}. Recent works have explored using score-based guidance to steer generative trajectories toward desired regions \citep{sabour2025test, yuan2024reward}. The relationship between score and velocity in flow models has been established through stochastic interpolants \citep{albergo2023stochastic}, showing that both encode equivalent information about the probability flow. We leverage this connection to derive a closed-form score expression from the velocity field, enabling drift modulation without auxiliary networks.

\section{Preliminaries}
\label{sec:prelim}

\paragraph{Reinforcement learning setup.}
We consider a Markov Decision Process (MDP) $\mathcal{M} = (\mathcal{S}, \mathcal{A}, \mathcal{P}, \gamma, \mathcal{R}, \mu)$, where $\mathcal{S}$ denotes the state space, $\mathcal{A} = \mathbb{R}^d$ the continuous action space, $\mathcal{P}: \mathcal{S} \times \mathcal{A} \to \Delta(\mathcal{S})$ the transition dynamics, $\gamma \in [0,1)$ the discount factor, $\mathcal{R}: \mathcal{S} \times \mathcal{A} \to \mathbb{R}$ the reward function, and $\mu$ the initial state distribution. The objective is to learn a policy $\pi_\theta(\mathbf{a} \mid \mathbf{s})$ that maximizes the expected discounted return:
\begin{equation}
J(\pi) = \mathbb{E}_{s_0 \sim \mu, a_t \sim \pi(\cdot \mid s_t), s_{t+1} \sim \mathcal{P}} \left[ \sum_{t=0}^{\infty} \gamma^t \mathcal{R}(s_t, a_t) \right].
\end{equation}
In offline-to-online RL, we first pre-train a policy via imitation learning on an offline dataset $\mathcal{D} = \{(s_i, a_i)\}_{i=1}^{|\mathcal{D}|}$, then fine-tune through online interaction to exceed demonstration performance.

\paragraph{Policy gradient and PPO.}
Policy gradient methods optimize $J(\pi)$ by estimating gradients with respect to policy parameters. The REINFORCE estimator \citep{williams1992simple} computes $\nabla_\theta J(\pi) = \mathbb{E}_{\pi}[\nabla_\theta \log \pi_\theta(\mathbf{a} \mid \mathbf{s}) \cdot A^\pi(\mathbf{s}, \mathbf{a})]$, where $A^\pi$ is the advantage function. This requires the policy to define a valid probability density $\pi_\theta(\mathbf{a} \mid \mathbf{s})$ over actions with tractable log-likelihood. Proximal Policy Optimization (PPO) \citep{schulman2017proximal} stabilizes training by constraining policy updates via a clipped surrogate:
\begin{equation}
\mathcal{L}^{\text{PPO}}(\theta) = \mathbb{E} \left[ \min \left( r_t(\theta) A_t, \, \text{clip}(r_t(\theta), 1-\epsilon, 1+\epsilon) A_t \right) \right],
\end{equation}
where $r_t(\theta) = \pi_\theta(\mathbf{a}_t \mid \mathbf{s}_t) / \pi_{\theta_{\text{old}}}(\mathbf{a}_t \mid \mathbf{s}_t)$ is the probability ratio. PPO is widely used for fine-tuning generative policies due to its stability and sample efficiency.

\paragraph{Flow Matching.}
Flow Matching (FM) \citep{lipman2022flow} learns a time-dependent velocity field $\mathbf{v}_\theta: \mathbb{R}^d \times [0,1] \to \mathbb{R}^d$ that transforms noise samples $\mathbf{a}_0 \sim \mathcal{N}(\mathbf{0}, \mathbf{I})$ into data samples $\mathbf{a}_1 \sim p_{\text{data}}$ through a continuous-time process. The linear interpolation between noise and data is given by $\mathbf{a}_t = (1-t)\mathbf{a}_0 + t\mathbf{a}_1$ for $t \in [0,1]$. The model learns to approximate this process by solving the ordinary differential equation (ODE) $\mathrm{d}\mathbf{a}_t = \mathbf{v}_\theta(\mathbf{a}_t, t)\mathrm{d}t$, and is trained using the flow matching objective:
\begin{equation}
\mathcal{L}_{\text{FM}}(\theta) = \mathbb{E}_{t, \mathbf{a}_0, \mathbf{a}_1} \left[ \left\| \mathbf{v}_\theta(\mathbf{a}_t, t) - (\mathbf{a}_1 - \mathbf{a}_0) \right\|_2^2 \right],
\end{equation}
where $t \sim \mathcal{U}[0,1]$, $\mathbf{a}_0 \sim \mathcal{N}(\mathbf{0}, \mathbf{I})$, and $\mathbf{a}_1 \sim p_{\text{data}}$. For policy learning, the velocity field is conditioned on observations: $\mathbf{v}_\theta(\mathbf{a}_t, t, \mathbf{s})$.

\section{Method}

We present ScoRe-Flow, a method for RL fine-tuning of Flow Matching policies that achieves decoupled control over the mean and variance of stochastic transitions. We first derive a closed-form score expression from the velocity field (Section~\ref{sec:score_derivation}), then introduce ScoRe-Flow which combines score-based drift modulation with learned variance prediction (Section~\ref{sec:scoreflow}).

\subsection{Closed-Form Score from Velocity}
\label{sec:score_derivation}

As discussed in Section~\ref{sec:prelim}, converting the deterministic FM dynamics to an SDE enables tractable likelihood computation for policy gradient methods. The general SDE formulation $\mathrm{d}\mathbf{a}_t = \boldsymbol{\mu}(\mathbf{a}_t, t)\mathrm{d}t + \sigma(\mathbf{a}_t, t)\mathrm{d}\mathbf{W}_t$ provides flexibility in designing both the drift $\boldsymbol{\mu}$ and diffusion $\sigma$. Existing methods such as ReinFlow \citep{zhang2025reinflow} set $\boldsymbol{\mu} = \mathbf{v}_\theta$ (the pre-trained velocity) and learn only $\sigma$. However, from a kinematic perspective, the discrete update $\mathbf{a}_{k+1} = \mathbf{a}_k + \mathbf{v}\Delta t + \sigma\epsilon$ reveals that the noise term $\sigma\epsilon$ only perturbs the current position without modulating the dynamics of the flow itself. This motivates us to explore whether modulating the drift can provide additional benefits.

\paragraph{Score function.}
To enable drift modulation without training additional networks, we leverage the \emph{score function}, i.e., the gradient of the log-density of intermediate marginals \citep{song2020score}. Let $\rho_t(\mathbf{a})$ denote the marginal density of the flow at time $t$. The score function is defined as
\begin{equation}
\mathbf{s}_t(\mathbf{a}) \triangleq \nabla_{\mathbf{a}}\log\rho_t(\mathbf{a}),
\label{eq:score_def}
\end{equation}
which points toward regions of higher probability density in the intermediate marginal, providing a natural direction for steering transitions toward the data manifold.

\paragraph{Derivation.}
For the linear FM path $\mathbf{a}_t = (1-t)\mathbf{a}_0 + t\mathbf{a}_1$ with $\mathbf{a}_0 \sim \mathcal{N}(\mathbf{0}, \mathbf{I})$ and $\mathbf{a}_1 \sim p_{\text{data}}$, the intermediate marginal $\rho_t$ is a convolution of the data distribution with Gaussian noise of variance $(1-t)^2\mathbf{I}$. By analyzing the relationship between the optimal velocity and the conditional expectation, we derive a closed-form expression for the score in terms of the velocity field (see Appendix~\ref{app:score_derivation} for the complete derivation):
\begin{equation}
\mathbf{s}_t(\mathbf{a}) = \frac{t\,\mathbf{v}_\theta(t,\mathbf{a},\mathbf{s})-\mathbf{a}}{1-t}.
\label{eq:score_closed_form}
\end{equation}

This \emph{score-velocity duality} is established in the stochastic interpolant literature \citep{albergo2023stochastic, albergo2025stochastic}, where the score and velocity are shown to encode equivalent information about the underlying probability flow (see also \citet{lipman2022flow}, Eq.~4.78). Our contribution is not the identity itself, but the observation that it enables practical drift modulation for RL fine-tuning of flow policies: the score can be computed directly from the pre-trained velocity field via a simple algebraic transformation, requiring zero additional parameters and negligible computational overhead, enabling decoupled control over the mean and variance of stochastic transitions.

Consistent with the boundary divergence issues noted in Elucidating Diffusion Models \citep{karras2022elucidating}, a critical practical consideration is that the score magnitude diverges as $|\mathbf{s}_t(\mathbf{a})| = O((1-t)^{-1})$ near $t=1$. Therefore, the coefficient multiplying the score in the drift must decay as $O(1-t)$ to keep the score-corrected drift bounded. In ScoRe-Flow, we enforce this property via a hard time-decay constraint on the score scheduler:
\begin{equation}
\alpha_\psi^{\text{scaled}}(t) \triangleq (1-t)\,\alpha_\psi(t),
\label{eq:alpha_scaled}
\end{equation}
so that $\alpha_\psi^{\text{scaled}} \to 0$ as $t \to 1$ and the score-modulated drift remains numerically stable.

%==============================================================================
\subsection{ScoRe-Flow: Decoupled Mean and Variance Control}
\label{sec:scoreflow}
%==============================================================================

\begin{algorithm}[t]
\caption{ScoRe-Flow: Action Sampling with Decoupled Mean-Variance Control}
\label{alg:score_flow}
\begin{algorithmic}[1]
\STATE \textbf{Input:} Observation $\mathbf{s}$, pre-trained velocity $\mathbf{v}_\theta$, learnable networks $\alpha_\psi, \sigma_\phi$, steps $K$
\STATE Sample initial noise: $\mathbf{a} \sim \mathcal{N}(\mathbf{0}, \mathbf{I})$
\STATE Initialize log-probability: $\log p \leftarrow 0$
\FOR{$k = 0, 1, \ldots, K-1$}
    \STATE Compute time step: $t \leftarrow k / K$, $\Delta t \leftarrow 1/K$
    \STATE Evaluate velocity field: $\mathbf{v} \leftarrow \mathbf{v}_\theta(\mathbf{a}, t, \mathbf{s})$
    \STATE Compute closed-form score: $\mathbf{s}_{\text{score}} \leftarrow (t \cdot \mathbf{v} - \mathbf{a}) / (1-t)$
    \STATE Predict learnable weights: $\alpha \leftarrow \alpha_\psi(t)$, $\sigma \leftarrow \sigma_\phi(\mathbf{a}, t, \mathbf{s})$
    \STATE Apply time-decay: $\alpha_{\text{scaled}} \leftarrow (1-t) \cdot \alpha$
    \STATE Compute transition mean: $\boldsymbol{\mu} \leftarrow \mathbf{a} + (\mathbf{v} + \alpha_{\text{scaled}} \cdot \mathbf{s}_{\text{score}}) \cdot \Delta t$
    \STATE Sample noise and update: $\boldsymbol{\epsilon} \sim \mathcal{N}(\mathbf{0}, \mathbf{I})$, $\mathbf{a} \leftarrow \boldsymbol{\mu} + \sigma \cdot \sqrt{\Delta t} \cdot \boldsymbol{\epsilon}$
    \STATE Accumulate log-probability: $\log p \leftarrow \log p + \log \mathcal{N}(\mathbf{a}; \boldsymbol{\mu}, \sigma^2 \Delta t \cdot \mathbf{I})$
\ENDFOR
\STATE \textbf{Return:} Final action $\mathbf{a}$, trajectory log-probability $\log p$
\end{algorithmic}
\end{algorithm}

With the closed-form score in hand, we now present ScoRe-Flow, which combines score-weighted drift modulation with learned variance prediction to achieve decoupled control over the first two moments of each transition kernel.

Noise-only methods learn $\sigma$ to control the magnitude of the exploration \citep{zhang2025reinflow}, but the noise term only perturbs the position at each step without modulating the underlying dynamics. Intuitively, this resembles adding random perturbations to a particle's location rather than adjusting its velocity. While effective for enabling exploration and likelihood computation, this approach does not leverage the geometric structure of the flow. In contrast, modulating the drift via the score, which points toward high-density regions consistent with the principles of Langevin Dynamics \citep{welling2011bayesian}, can provide directional guidance that complements variance control, improving stability when trajectories enter low-density regions and accelerating convergence when the pre-trained trajectory is already informative. However, standard score-based control rigidly couples the drift and diffusion terms (e.g., a drift of $\lambda(t)\mathbf{s}_t$ requires variance $\sqrt{2\lambda(t)}$ to maintain equilibrium), preventing independent adjustment of exploration magnitude and intensity required for effective RL fine-tuning.

ScoRe-Flow decouples drift and variance control by introducing two learnable components: a score scheduler $\alpha_\psi^{\text{scaled}}$ (with hard time decay from \cref{eq:alpha_scaled}) that modulates the score-based drift correction, and a variance predictor $\sigma_\phi$ that independently determines the exploration magnitude. The resulting SDE takes the form:
\begin{equation}
\begin{split}
\mathrm{d}\mathbf{a}_t = \Big[\mathbf{v}_{\theta}(t,\mathbf{a}_t,\mathbf{s}) + \alpha_\psi^{\text{scaled}}(t) \cdot \mathbf{s}_t(\mathbf{a}_t)\Big]\mathrm{d}t \\
+ \sigma_\phi(t,\mathbf{a}_t,\mathbf{s}) \, \mathrm{d}\mathbf{W}_t,
\end{split}
\label{eq:scoreflow_sde}
\end{equation}
where $\mathbf{v}_\theta$ is the FM velocity field (initialized from pre-training and jointly optimized during fine-tuning), $\mathbf{s}_t$ is the closed-form score from \cref{eq:score_closed_form}, $\alpha_\psi^{\text{scaled}}$ is the score scheduler with endpoint stabilization from \cref{eq:alpha_scaled}, and $\sigma_\phi: \mathbb{R}^d \times [0,1] \times \mathcal{S} \to \mathbb{R}^+$ is the learnable variance prediction network.

The key insight of ScoRe-Flow is the decomposition of distributional control into two complementary mechanisms. The score-modulated drift term $\alpha_\psi^{\text{scaled}} \cdot \mathbf{s}_t$ leverages the score to provide geometrically meaningful guidance toward higher-probability regions under the flow marginal. When the exploration trajectory wanders into low-density areas, the score acts as a corrective force steering it back toward the data manifold. The score scheduler $\alpha_\psi^{\text{scaled}}(t)$ adaptively determines the strength of this correction based on the time step, while satisfying the endpoint stability constraint via the $(1-t)$ decay. Meanwhile, the variance term $\sigma_\phi$ governs the magnitude of stochastic exploration independently of its direction. This decoupling enables the policy to, for example, maintain strong drift correction while reducing exploration noise as training progresses, or vice versa. Such flexibility is unattainable by noise-only or score-only methods.

\paragraph{Discrete-time formulation.}
For practical implementation, we discretize \cref{eq:scoreflow_sde} using the Euler-Maruyama scheme with $K$ integration steps. Let $t_k = k/K$ and $\Delta t = 1/K$ for $k = 0, 1, \ldots, K-1$. The discrete update rule is:
\begin{equation}
\mathbf{a}^{k+1} = \mathbf{a}^k + \Big[\mathbf{v}_\theta^k + (\alpha_\psi^{\text{scaled}})^k \cdot \mathbf{s}^k\Big]\Delta t + \sigma_\phi^k \sqrt{\Delta t}\,\boldsymbol{\epsilon}_k,
\label{eq:scoreflow_discrete}
\end{equation}
where $\boldsymbol{\epsilon}_k \sim \mathcal{N}(\mathbf{0}, \mathbf{I})$ is standard Gaussian noise, and we use the shorthand $\mathbf{v}_\theta^k = \mathbf{v}_\theta(t_k, \mathbf{a}^k, \mathbf{s})$, $(\alpha_\psi^{\text{scaled}})^k = (1-t_k)\alpha_\psi(t_k)$, $\mathbf{s}^k = \mathbf{s}_{t_k}(\mathbf{a}^k)$, and $\sigma_\phi^k = \sigma_\phi(t_k, \mathbf{a}^k, \mathbf{s})$.

\paragraph{Likelihood computation.}
Each transition follows a Gaussian distribution with mean $\boldsymbol{\mu}^k = \mathbf{a}^k + [\mathbf{v}_\theta^k + (\alpha_\psi^{\text{scaled}})^k \mathbf{s}^k]\Delta t$ and variance $(\sigma_\phi^k)^2 \Delta t$:
\begin{equation}
p(\mathbf{a}^{k+1} \mid \mathbf{a}^k, \mathbf{s}) = \mathcal{N}\Big(\mathbf{a}^{k+1}; \boldsymbol{\mu}^k, (\sigma_\phi^k)^2 \Delta t \, \mathbf{I}\Big).
\label{eq:transition_density}
\end{equation}
The trajectory log-likelihood factorizes as:
\begin{equation}
\log \pi(\mathbf{a}_1 \mid \mathbf{s}) = \sum_{k=0}^{K-1} \log p(\mathbf{a}^{k+1} \mid \mathbf{a}^k, \mathbf{s}),
\label{eq:trajectory_loglik}
\end{equation}
which can be computed in closed form, enabling direct application of PPO for policy optimization.

% \paragraph{Implementation.}
% Both $\alpha_\psi$ and $\sigma_\phi$ are implemented as lightweight multi-layer perceptrons (MLPs) that share a common encoder. The encoder takes as input the concatenation of the current action $\mathbf{a}_t$, a sinusoidal positional embedding of the time step $t$, and the observation $\mathbf{s}$ (or its encoded representation for image observations). Two separate output heads produce the scalar $\alpha_\psi$ and $\sigma_\phi$ values. This shared architecture ensures that both components condition on the same state representation while adding minimal computational overhead to the base policy evaluation.

% During RL fine-tuning, we freeze the pre-trained velocity field $\mathbf{v}_\theta$ and optimize only the parameters $\psi$ and $\phi$ of the drift modulation and variance networks using PPO with the trajectory log-likelihood from \cref{eq:trajectory_loglik}. The polynomial schedule $\lambda(t) = \lambda_{\max}(1-t)^{\gamma}$ with $\gamma \geq 1$ ensures that the score contribution is strongest in the early stages of the flow (where the marginal is closer to the prior) and diminishes toward the endpoint (where the score diverges). We use $\gamma = 2$ in all experiments unless otherwise specified. Complete architecture specifications and hyperparameter settings are provided in Appendix~\ref{app:architecture} and~\ref{app:hyperparams}.

\paragraph{Implementation.}
The drift modulation weight $\alpha_\psi(t)$ is implemented as a lightweight MLP that takes only the scalar time step $t$ as input. The variance predictor $\sigma_\phi$ is an independent MLP that takes the concatenation of the current action $\mathbf{a}_t$, time step $t$, and observation $\mathbf{s}$ (or its encoded representation for image observations).
During RL fine-tuning, we adopt a full parameter fine-tuning strategy. Instead of freezing the base model, we jointly optimize the pre-trained velocity field $\mathbf{v}_\theta$ alongside the parameters $\psi$ and $\phi$ of the drift modulation and variance networks using PPO, maximizing the trajectory log-likelihood defined in Eq.~\ref{eq:trajectory_loglik}.
The hard time-decay constraint in \cref{eq:alpha_scaled} ensures that the score correction is strongest in the early stages of the flow and diminishes toward the endpoint, preventing instability from the $(1-t)^{-1}$ scaling of the score. Complete architecture specifications and hyperparameter settings are provided in Appendix~\ref{app:architecture} and~\ref{app:hyperparams}.

\section{Experiments}

We evaluate ScoRe-Flow on continuous control benchmarks spanning locomotion and manipulation to answer: (i) Does combining score-based mean control with variance prediction outperform either component alone? (ii) How does ScoRe-Flow compare with state-of-the-art methods?

\subsection{Environments}

We evaluate on three benchmark suites.

\textbf{D4RL Locomotion} \citep{fu2020d4rl} comprises OpenAI Gym MuJoCo tasks including Hopper-v2, Walker2d-v2, Ant-v0, and Humanoid-v3, where we use the medium-expert datasets for pre-training. These tasks provide dense rewards and test the ability to learn efficient locomotion gaits beyond suboptimal demonstrations. 

\textbf{Robomimic} \citep{mandlekar2021matters} consists of visual manipulation tasks including PickPlaceCan (single-arm pick-and-place), Square (precise peg insertion) and Transport (bimanual coordination), with 96$\times$96 RGB image observations and proficient-human (ph) demonstrations for pre-training.

\textbf{Franka Kitchen} \citep{gupta2019relay} is a benchmark environment for sequential multi-task manipulation, which includes tasks such as opening a microwave, repositioning a kettle, activating a light switch, and sliding a cabinet door using a Franka Panda robotic arm. We perform evaluations on the complete, mixed, and partial dataset variants under a sparse reward formulation in which a unit reward is granted upon successful completion of each subtask.

\begin{figure*}[!t]
    \centering
    \begin{subfigure}[b]{0.23\textwidth}
        \centering
        \includegraphics[width=\textwidth]{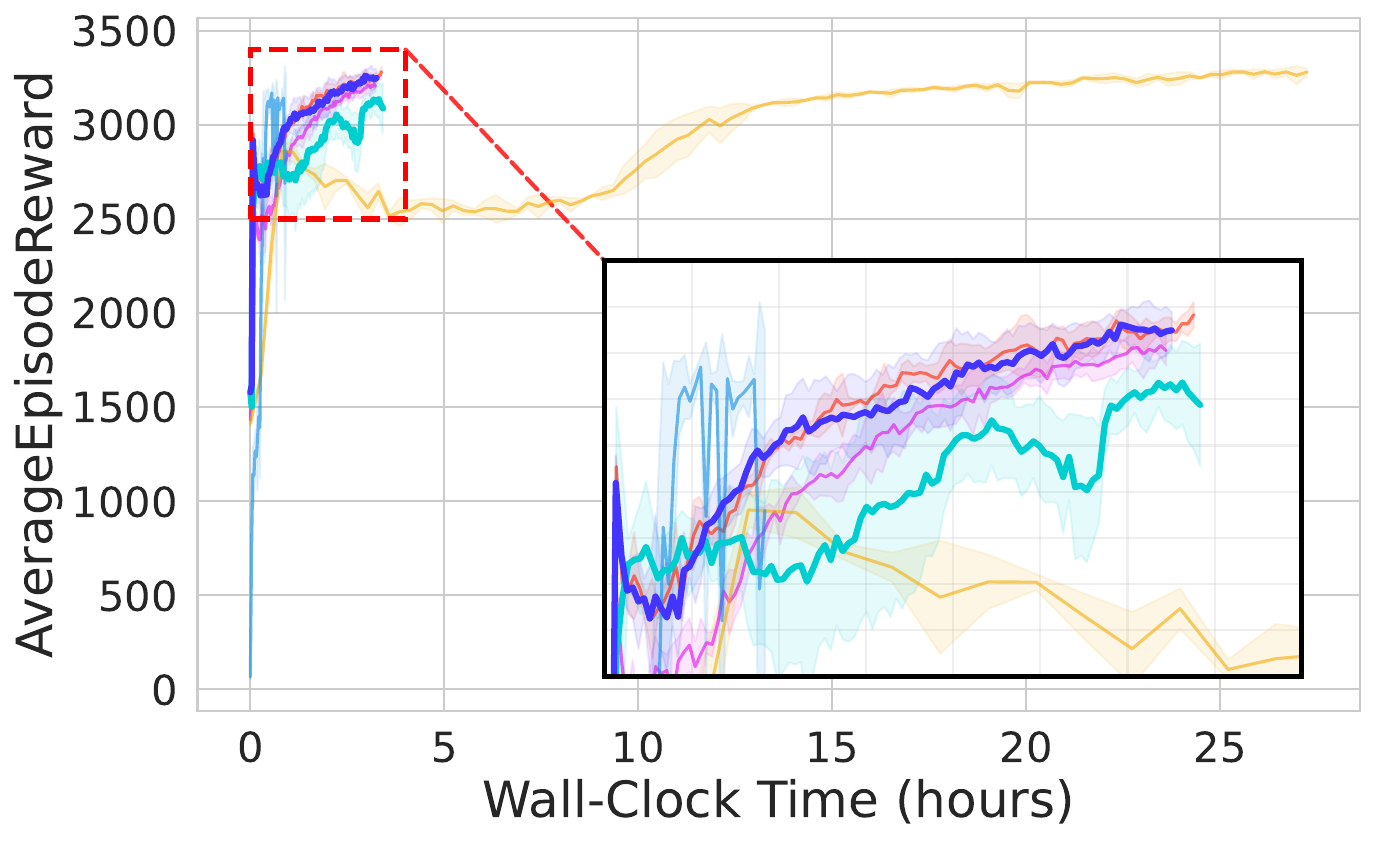}
        \caption{Hopper-v2}
    \end{subfigure}
    \hfill
    \begin{subfigure}[b]{0.23\textwidth}
        \centering
        \includegraphics[width=\textwidth]{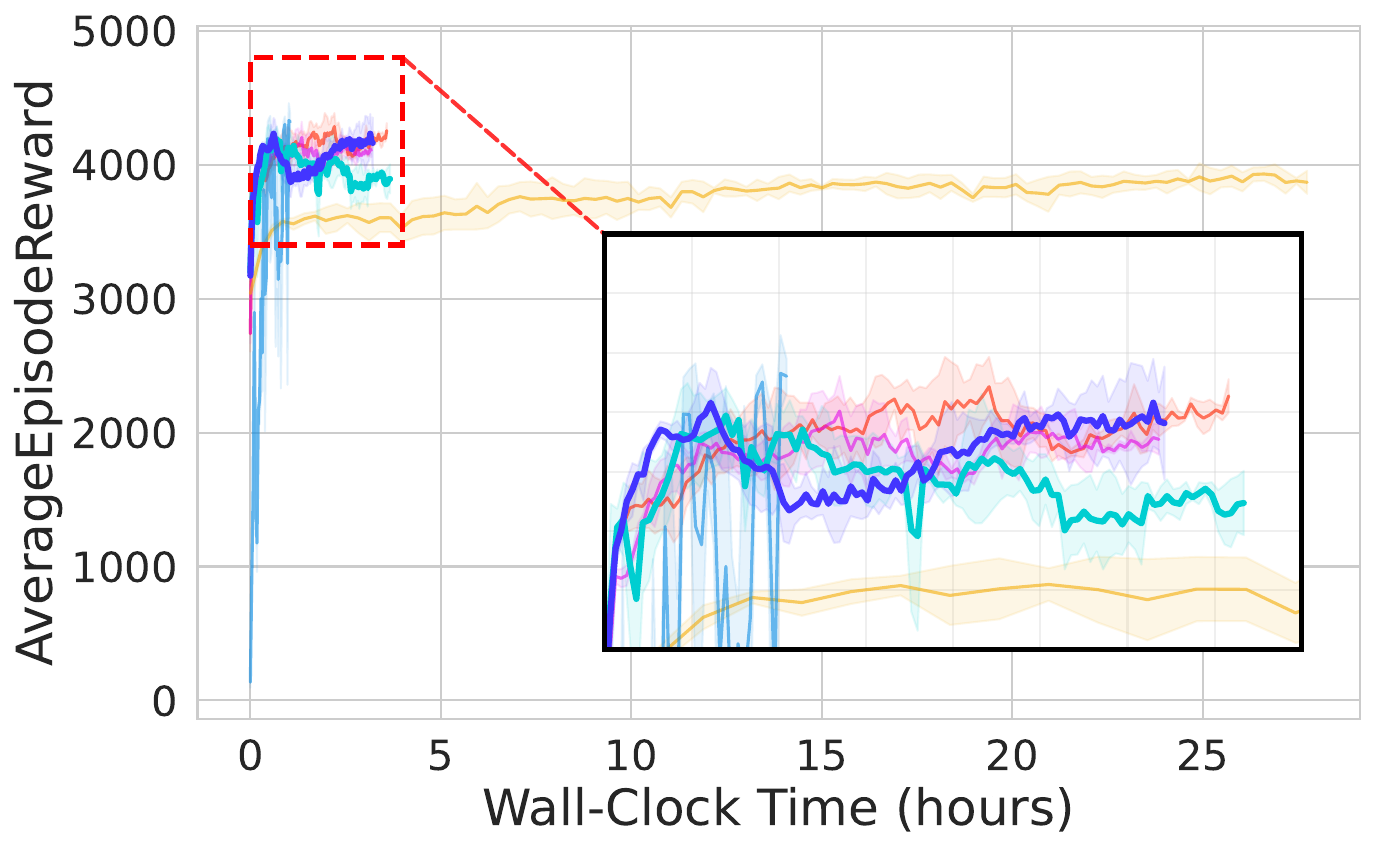}
        \caption{Walker2d-v2}
    \end{subfigure}
    \hfill
    \begin{subfigure}[b]{0.23\textwidth}
        \centering
        \includegraphics[width=\textwidth]{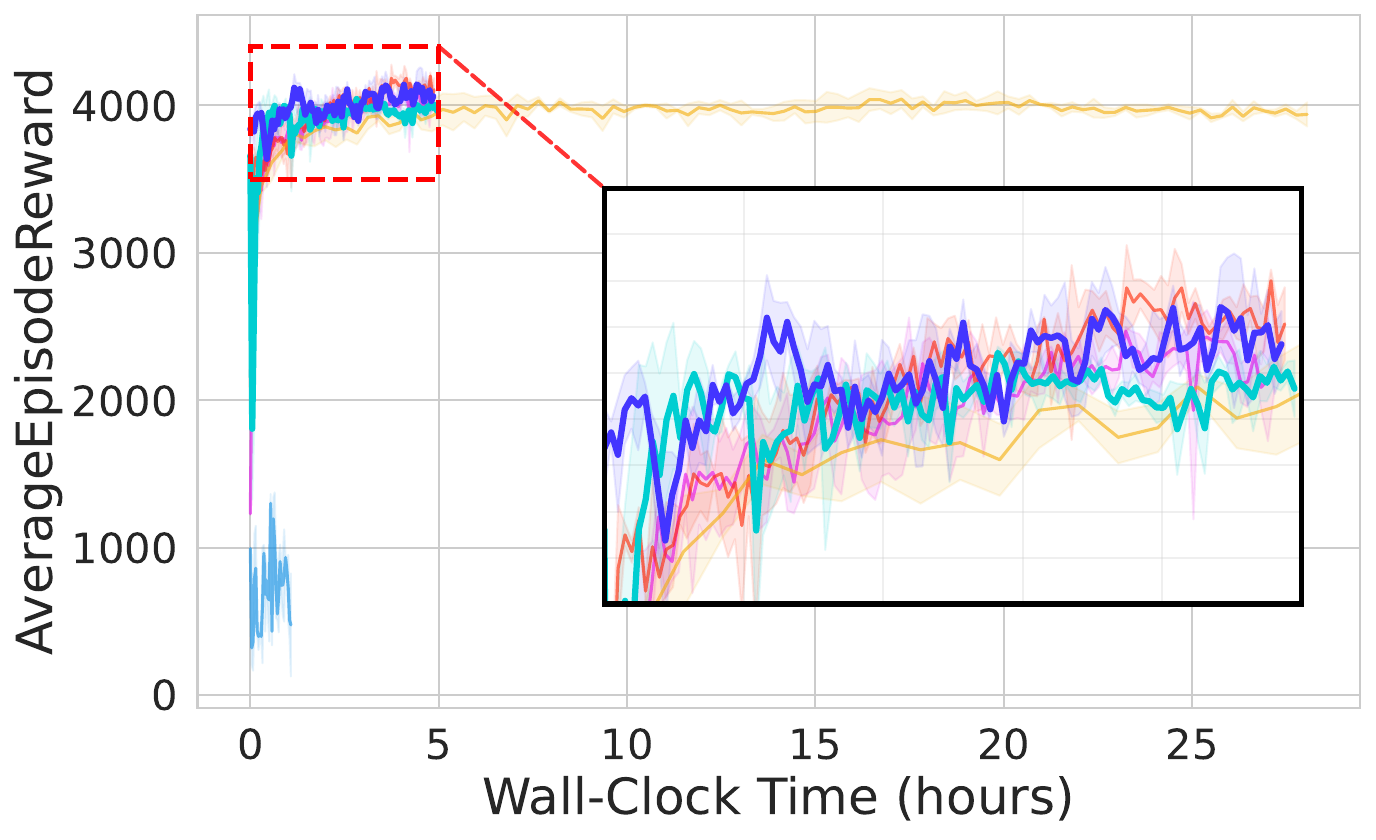}
        \caption{Ant-v0}
    \end{subfigure}
    \hfill
    \begin{subfigure}[b]{0.23\textwidth}
        \centering
        \includegraphics[width=\textwidth]{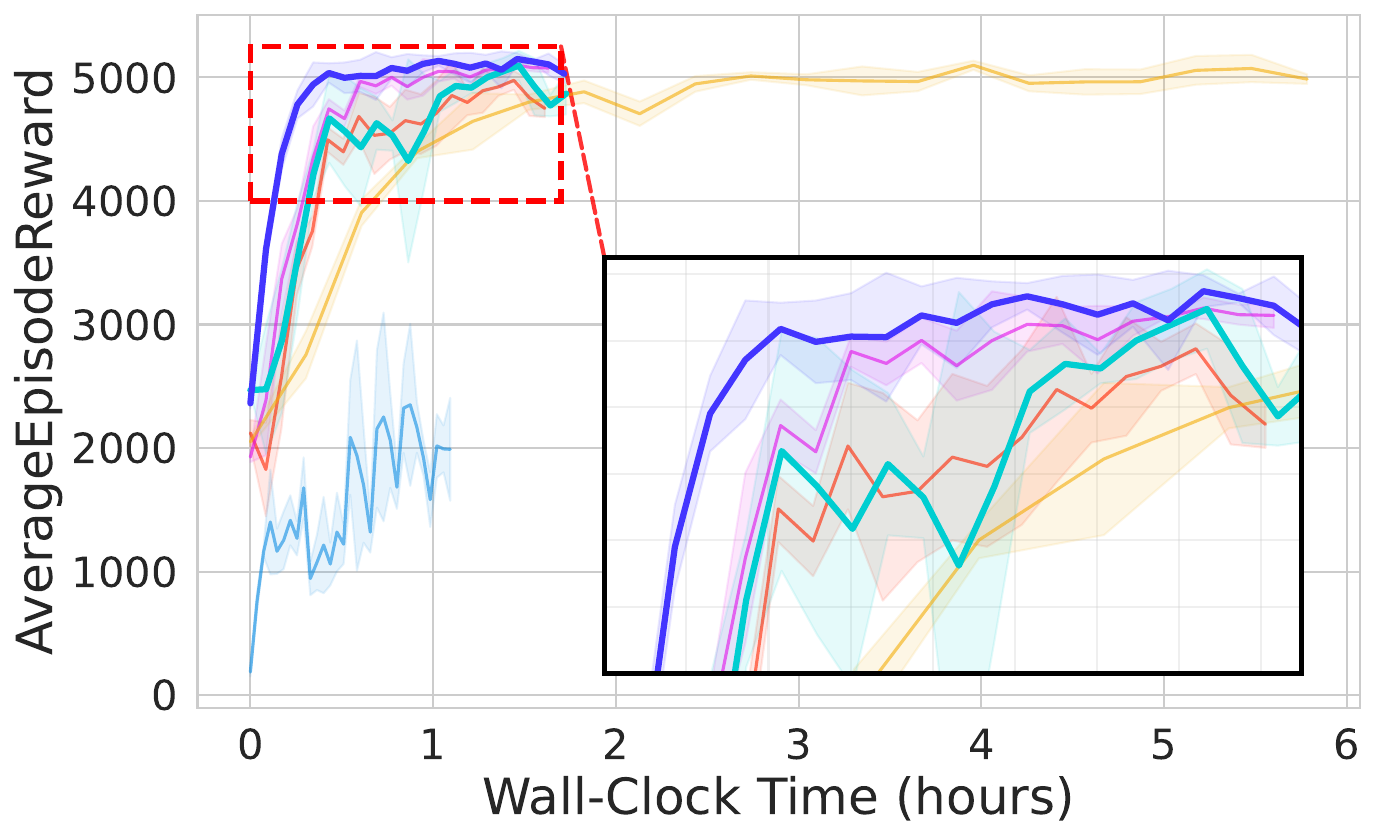}
        \caption{Humanoid-v3}
    \end{subfigure}
    
    \includegraphics[width=0.7\textwidth]{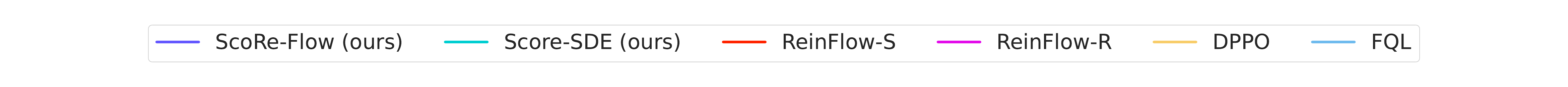}
    \caption{Learning curves on D4RL locomotion tasks. Dashed lines indicate the behavior cloning level.}
    \label{fig:d4rl}
\end{figure*}

\begin{figure*}[!t]
    \centering
    \begin{subfigure}[b]{0.32\textwidth}
        \centering
        \includegraphics[width=\textwidth]{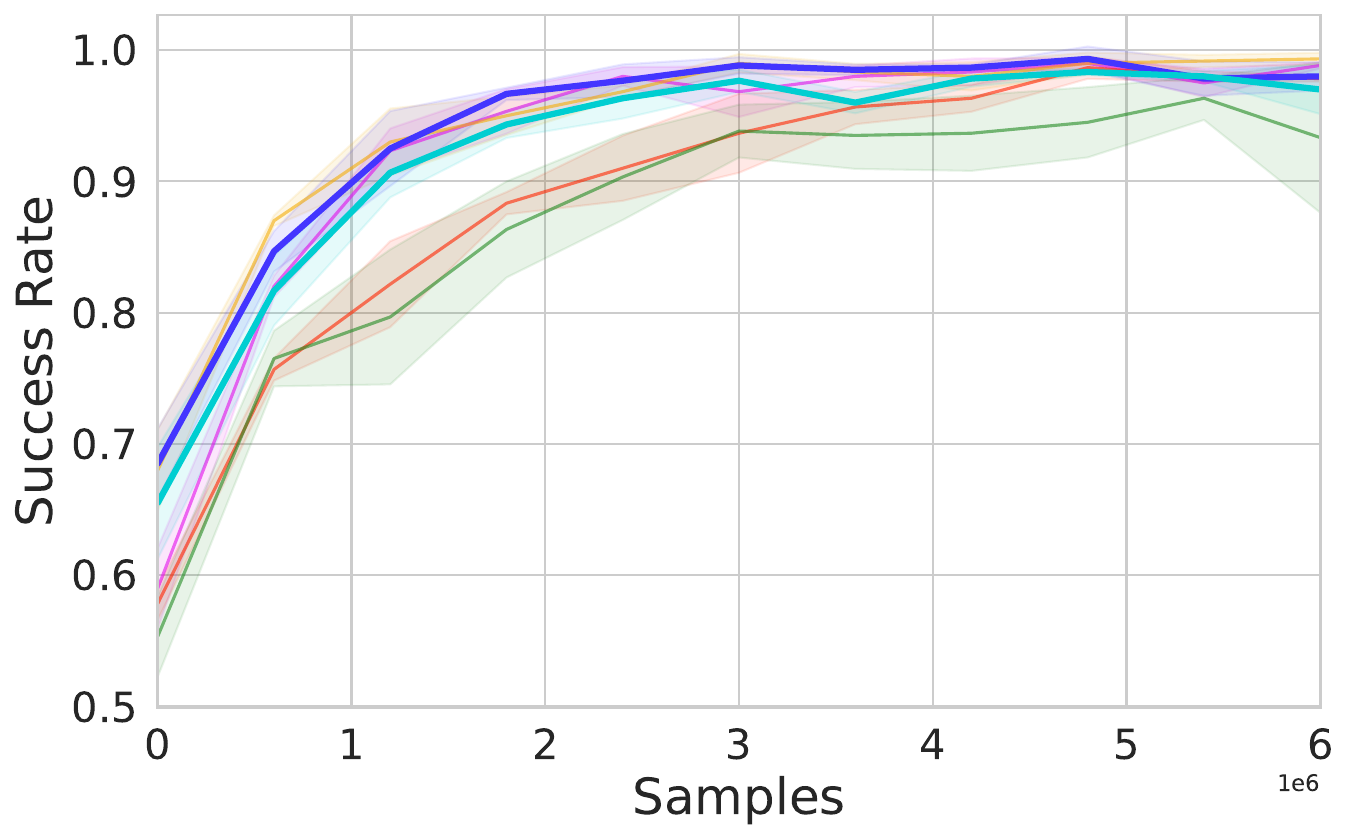}
        \caption{PickPlaceCan}
    \end{subfigure}
    \hfill
    \begin{subfigure}[b]{0.32\textwidth}
        \centering
        \includegraphics[width=\textwidth]{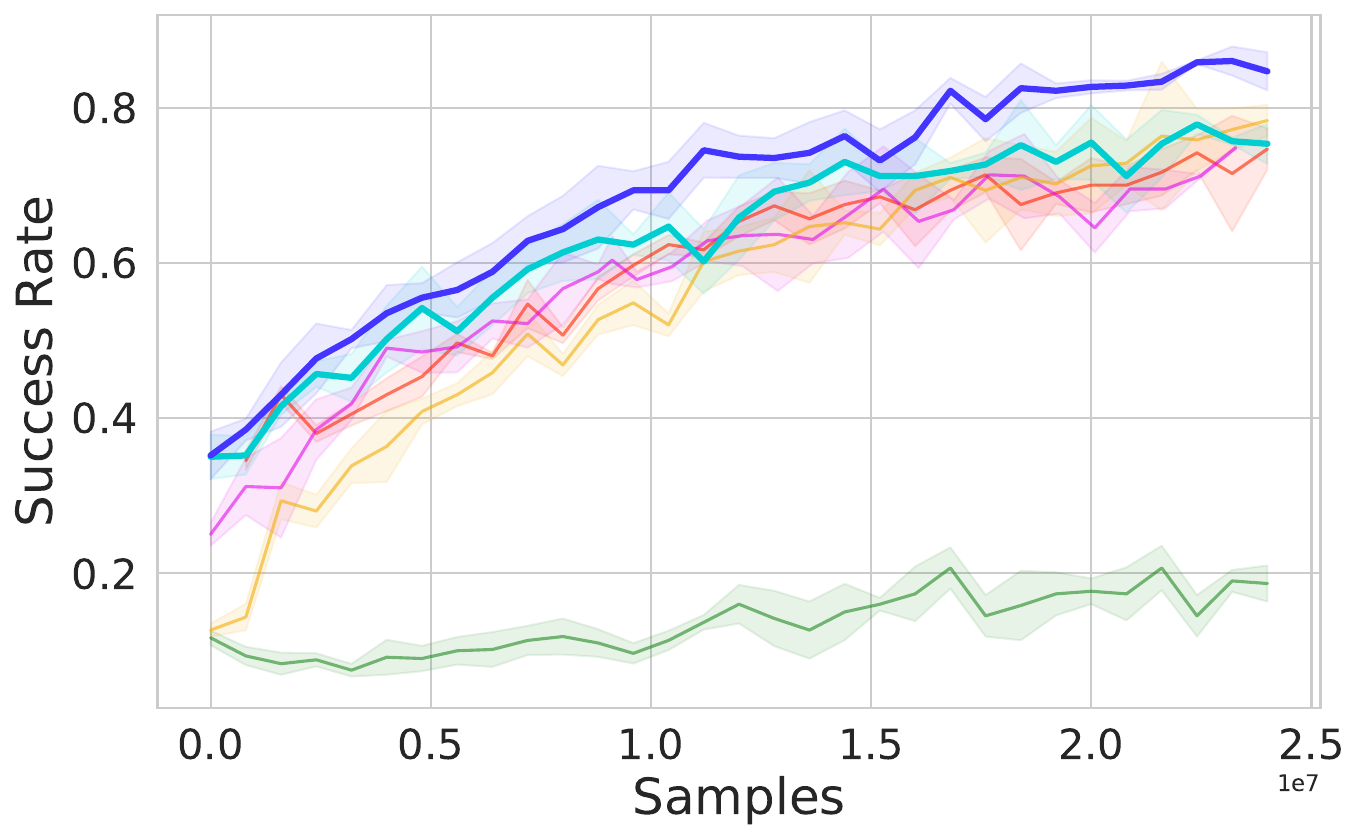}
        \caption{Square}
    \end{subfigure}
    \hfill
    \begin{subfigure}[b]{0.32\textwidth}
        \centering
        \includegraphics[width=\textwidth]{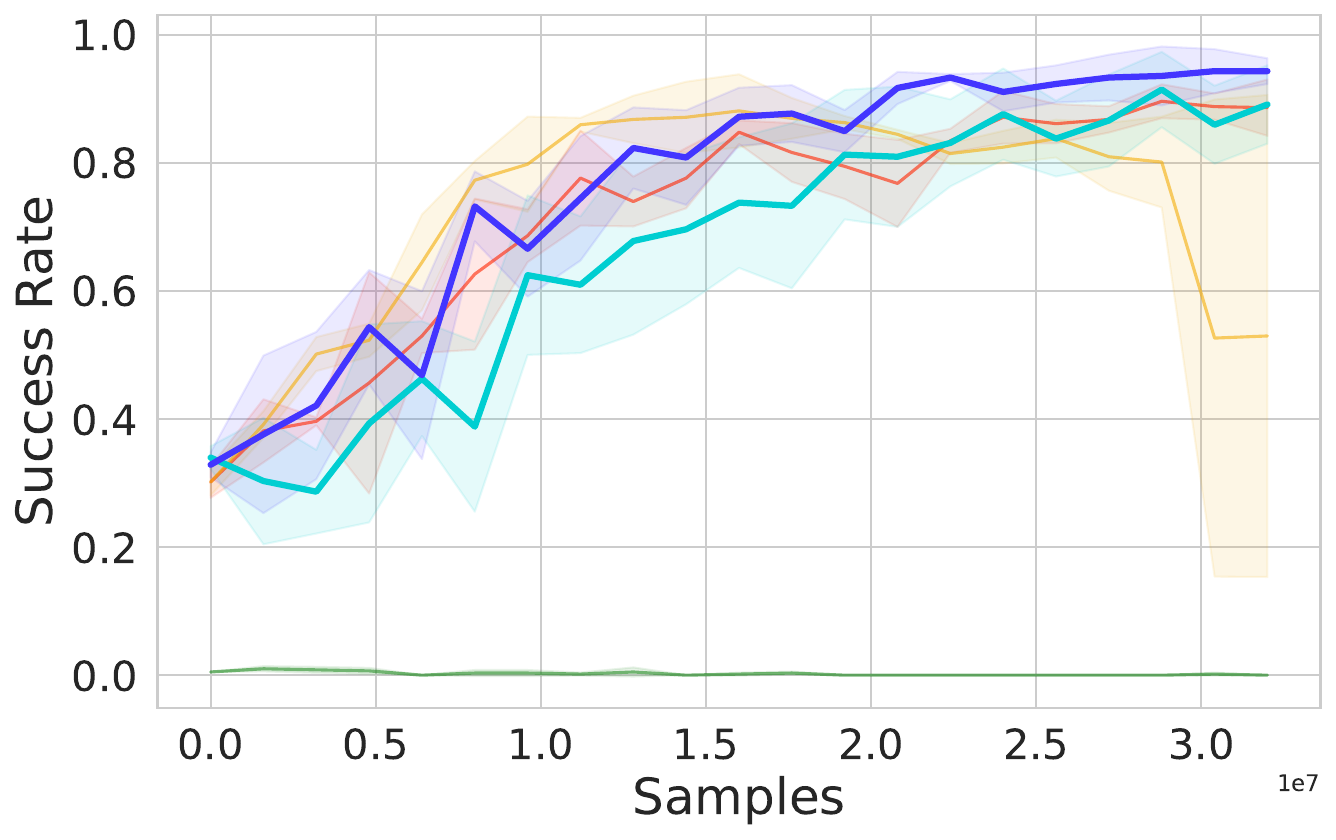}
        \caption{Transport}
    \end{subfigure}
    
    \includegraphics[width=0.7\textwidth]{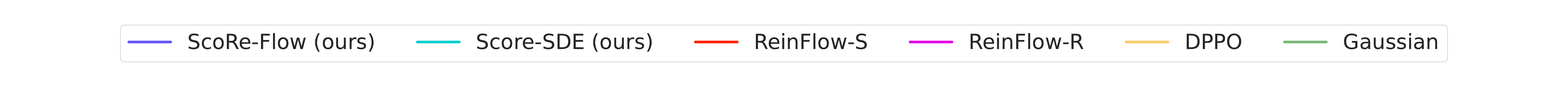}
    \caption{Learning curves on Robomimic visual manipulation tasks.}
    \label{fig:robomimic}
\end{figure*}

\subsection{Baselines}

We compare ScoRe-Flow against the following methods.

\textbf{ReinFlow} \citep{zhang2025reinflow} introduces learnable variance prediction for flow policy RL. We evaluate two variants: \textbf{ReinFlow-R} uses Rectified Flow as the base policy with 4 denoising steps, representing the variance-only control paradigm; \textbf{ReinFlow-S} uses Shortcut Model as the base policy with 1 to 4 denoising steps and learned step-skipping for improved efficiency.

\textbf{DPPO} \citep{ren2024diffusion} applies Diffusion Policy \citep{chi2023diffusion} with PPO fine-tuning, using DDIM sampling with 50 to 100 denoising steps at higher computational cost but achieving strong performance on diffusion policies.

\textbf{FQL} \citep{park2025flow} implements Flow Q-Learning with offline training and distillation, lacking online exploration during the fine-tuning phase.

\paragraph{Ablation Method.}

\textbf{Score-SDE} is a simplified variant that uses only score-based mean control without learned variance prediction. In this ablation, the same schedule $\lambda(t)$ determines both the drift correction strength and the diffusion magnitude. Specifically, it follows the SDE:
\begin{equation}
\mathrm{d}\mathbf{a}_t = \Big[\mathbf{v}_\theta + \lambda(t) \cdot \mathbf{s}_t(\mathbf{a}_t)\Big]\mathrm{d}t + \sqrt{2\lambda(t)} \, \mathrm{d}\mathbf{W}_t,
\label{eq:score_sde}
\end{equation}
where the diffusion coefficient is coupled to the drift schedule as $\sigma(t) = \sqrt{2\lambda(t)}$. This formulation corresponds to a time-reversal of the forward noising process. By comparing Score-SDE to ScoRe-Flow, we isolate the contribution of decoupled variance prediction.

% \paragraph{Pre-trained Policies.}

% We use Rectified Flow policies as the base model for fair comparison. All pre-trained policies are trained using the Flow Matching objective on the respective demonstration datasets. For D4RL, we use 4 denoising steps and action chunk size 4. For Robomimic, we use 1 to 4 denoising steps depending on task complexity. For Franka Kitchen, we use 4 denoising steps with action chunk size 4 and condition stacking 1.

\paragraph{Pre-trained Policies.}
To ensure a rigorous comparison, we strictly align our pre-training setup with ReinFlow~\citep{zhang2025reinflow}. Specifically, we initialize our base policies using identical pre-trained weights and model architectures, adopting either Rectified Flow or Shortcut Models based on the optimal configuration reported in their work. All policies are trained using the standard Flow Matching objective on respective demonstration data.

\paragraph{Implementation Details.}
We adopt a task-dependent architecture for the pre-trained base policy, employing \textbf{Rectified Flow} for D4RL and simpler Robomimic tasks, and \textbf{Shortcut Models} for long-horizon tasks (Franka Kitchen, Transport) to maximize efficiency. 
For fine-tuning, we generally set the denoising steps $K=4$, with exceptions for specific tasks to balance performance and cost. The exploration process is governed by a learnable noise scheduler or a fixed hold strategy depending on the domain.
Comprehensive hyperparameters, including network architectures and detailed optimization settings, are provided in Tables~\ref{tab:shared_params_detailed} to \ref{tab:robomimic_params_combined}.

% For ScoRe-Flow, we use $K = 4$ integration steps with schedule $\lambda(t) = \lambda_{\max}(1-t)^\gamma$ where $\lambda_{\max} \in [0.1, 0.3]$ and $\gamma = 1.5$. The networks $\alpha_\psi$ and $\sigma_\phi$ share a 2-layer MLP encoder with 256 hidden units. We employ PPO as the RL algorithm with clipping ratio $\epsilon = 0.01$ and GAE $\lambda = 0.95$. For D4RL, we use 40 parallel environments with 1000 training iterations. For Robomimic, we use 50 parallel environments with gradient accumulation over 15 steps and image augmentation (RandomShift). For Franka Kitchen, we use 40 parallel environments with 300 training iterations. Full hyperparameters are provided in Appendix~\ref{app:hyperparams}.

\begin{figure*}[!t]
    \centering
    \begin{subfigure}[b]{0.32\textwidth}
        \centering
        \includegraphics[width=\textwidth]{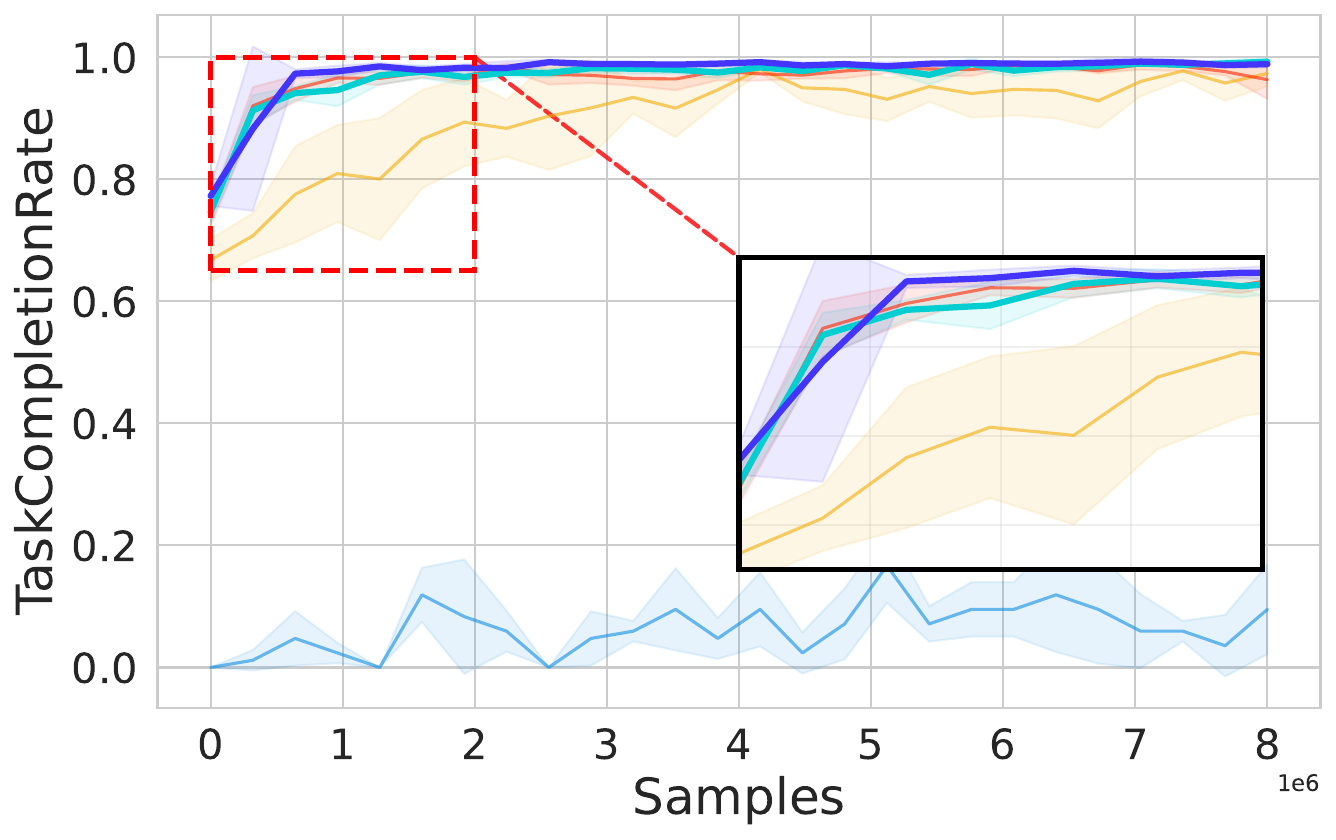}
        \caption{Complete}
    \end{subfigure}
    \hfill
    \begin{subfigure}[b]{0.32\textwidth}
        \centering
        \includegraphics[width=\textwidth]{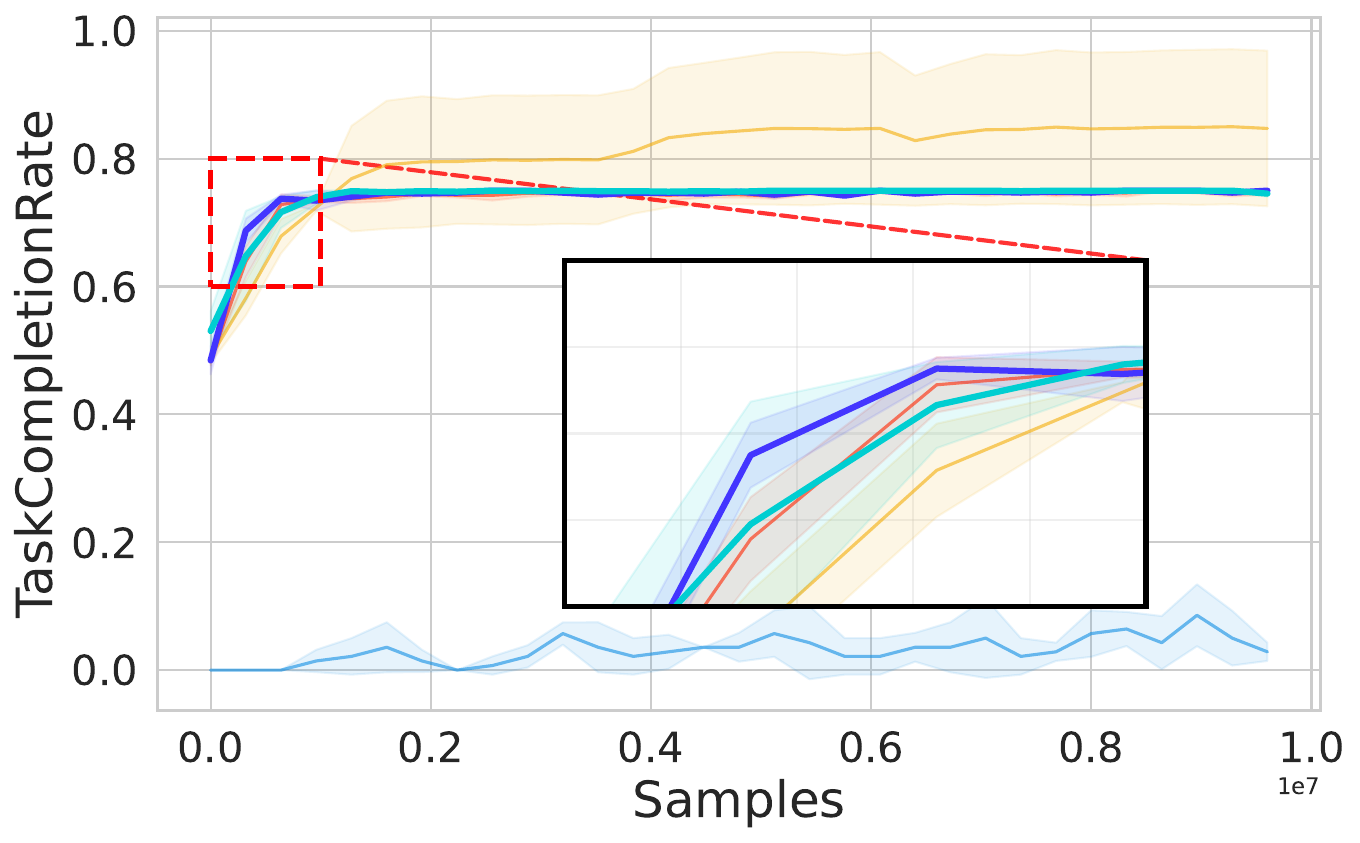}
        \caption{Mixed}
    \end{subfigure}
    \hfill
    \begin{subfigure}[b]{0.32\textwidth}
        \centering
        \includegraphics[width=\textwidth]{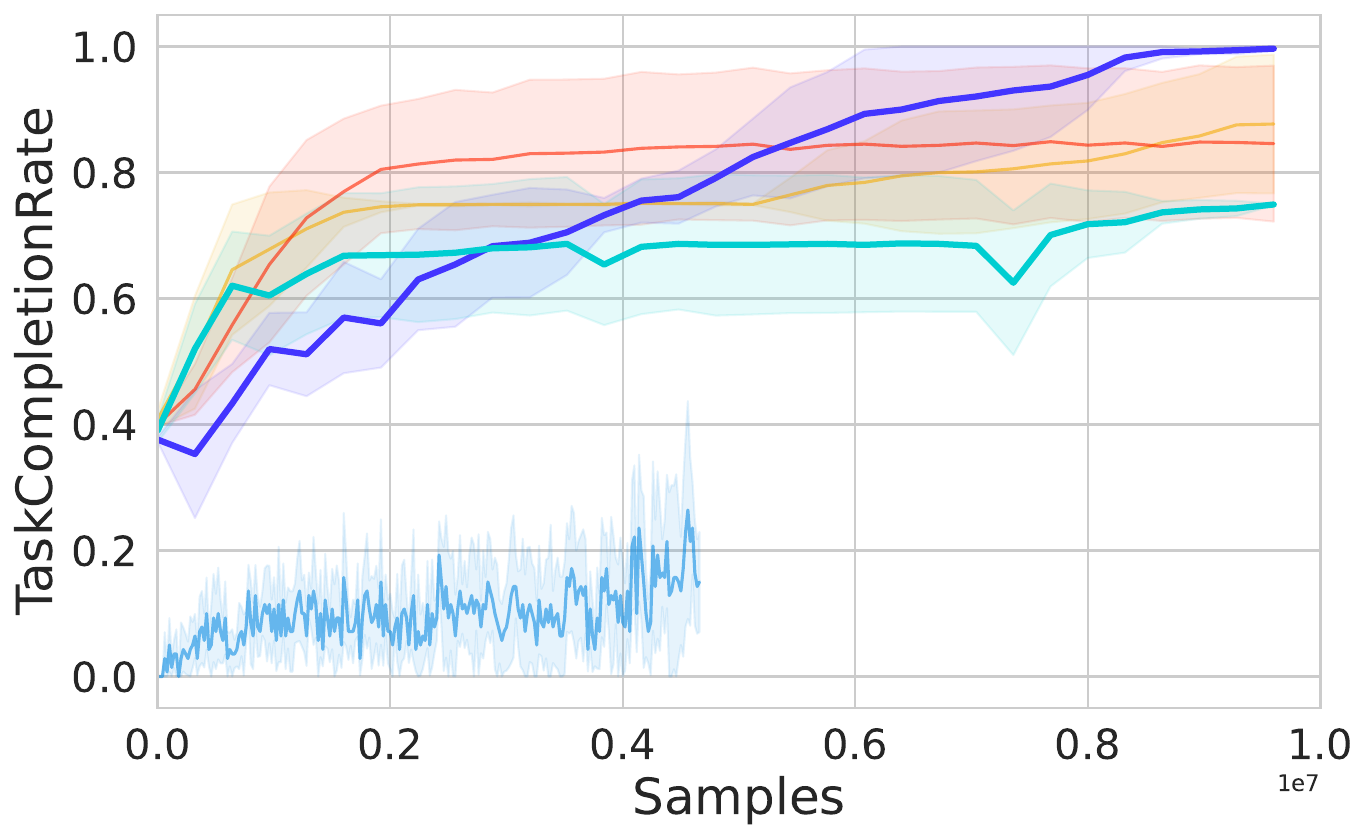}
        \caption{Partial}
    \end{subfigure}
    
    \includegraphics[width=0.7\textwidth]{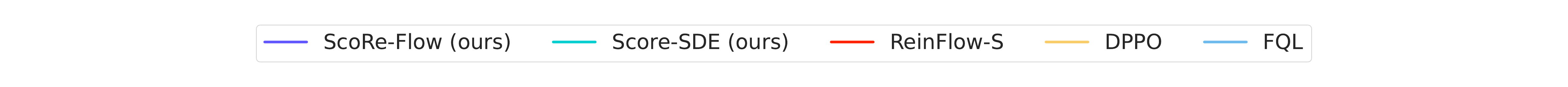}
    \caption{Learning curves on Franka Kitchen multi-task benchmark.}
    \label{fig:kitchen}
\end{figure*}

\begin{figure*}[!t]
    \centering
    \includegraphics[width=2\columnwidth]{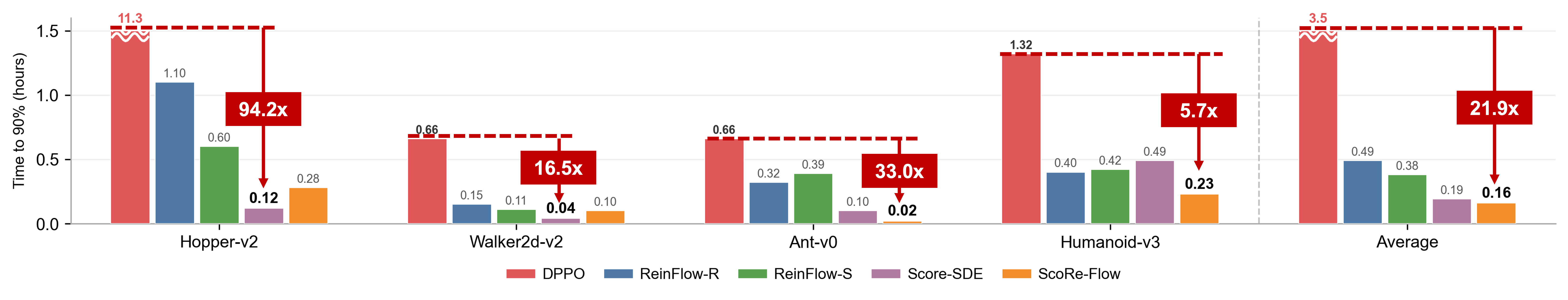}
    \caption{Wall-clock time comparison on D4RL locomotion tasks. The large speedup over DPPO (up to 94.2$\times$ on Hopper-v2, 21.9$\times$ average) is primarily a structural advantage of flow-based methods requiring fewer denoising steps. The algorithmic contribution is the $2.4\times$ faster convergence over the flow-based baseline ReinFlow at matched $K=4$ steps.}
    \label{fig:time_comparison}
\end{figure*}

\begin{table*}[!t]
\centering
\caption{Performance comparison across all benchmarks. We report final performance (mean $\pm$ std over 3 seeds). The BC (pretrain) column indicates the baseline performance of the behavior cloning policy before fine-tuning. \textbf{Bold}: best, \underline{underline}: second best.}
\label{tab:main_results}
\small
\setlength{\tabcolsep}{8pt} % 稍微减小列间距以容纳新增的一列
\renewcommand{\arraystretch}{1.1}
\begin{tabular}{lcccccc}
\toprule
Task & BC (pretrain) & DPPO & ReinFlow-R & ReinFlow-S & Score-SDE & \textbf{ScoRe-Flow (Ours)} \\
\midrule
\multicolumn{7}{c}{\textit{D4RL Locomotion (Episode Reward $\uparrow$)}} \\
\midrule
Hopper-v2   & 1578$\pm$28& \textbf{3275$\pm$20} & 3204$\pm$36 & \underline{3253$\pm$24} & 3118$\pm$87 & 3233$\pm$39 \\
Walker2d-v2 & 3185$\pm$38& 3896$\pm$62 & 4092$\pm$11 & \textbf{4202$\pm$26} & 3900$\pm$57 & \underline{4172$\pm$154} \\
Ant-v0      & 3750$\pm$118& 3951$\pm$24 & 4021$\pm$25 & \textbf{4119$\pm$44} & 3980$\pm$29 & \underline{4088$\pm$95} \\
Humanoid-v3 & 2415$\pm$65& 5001$\pm$20 & \underline{5041$\pm$6} & 4801$\pm$134 & 4896$\pm$91 & \textbf{5100$\pm$47} \\
\rowcolor{gray!15} \textit{Avg.} & 2732 & 4031 & 4090 & \underline{4094} & 3974 & \textbf{4148} \\
\midrule
\multicolumn{7}{c}{\textit{Robomimic (Success Rate \% $\uparrow$)}} \\
\midrule
PickPlaceCan & 68.0$\pm$1.8& 96.5$\pm$0.2 & 95.6$\pm$0.1 & 91.7$\pm$0.5 & \underline{96.8$\pm$0.2} & \textbf{98.3$\pm$0.5} \\
Square       & 33.3$\pm$2.8& \underline{78.3$\pm$2.0} & 76.2$\pm$2.0 & 77.3$\pm$2.1 & 75.3$\pm$2.6 & \textbf{84.7$\pm$2.5} \\
Transport    & 27.0$\pm$3.6& 53.0$\pm$37.7 & N/A & 88.7$\pm$4.4 & \underline{89.2$\pm$6.1} & \textbf{94.4$\pm$2.0} \\
\rowcolor{gray!15} \textit{Avg.} & 42.8& 75.9 & N/A & 85.9 & \underline{87.1} & \textbf{92.5} \\
\midrule
\multicolumn{7}{c}{\textit{Franka Kitchen (Tasks Completed, max 4 $\uparrow$)}} \\
\midrule
Complete & 3.03$\pm$0.1& 3.8$\pm$0.1 & N/A & 3.9$\pm$0.0 & \textbf{4.0$\pm$0.0} & \textbf{4.0$\pm$0.0} \\
Mixed    & 2.04$\pm$0.1& \textbf{3.4$\pm$0.5} & N/A & 3.0$\pm$0.0 & 3.0$\pm$0.0 & 3.0$\pm$0.0 \\
Partial  & 1.53$\pm$0.1& 3.3$\pm$0.4& N/A & \underline{3.4$\pm$0.5} & 2.8$\pm$0.2 & \textbf{3.8$\pm$0.2} \\
\rowcolor{gray!15} \textit{Avg.} & 2.20& \underline{3.50} & N/A & 3.43 & 3.27 & \textbf{3.60} \\
\bottomrule
\end{tabular}
\end{table*}

% \twocolumn[{%
% \centering
% \begin{minipage}{0.30\textwidth}
%     \centering
%     \includegraphics[width=\textwidth]{figures/kitchen_kitchen-complete-v0_TaskCompletionRate_zoom.pdf}
    
%     {\small (a) Complete}
% \end{minipage}
% \hfill
% \begin{minipage}{0.30\textwidth}
%     \centering
%     \includegraphics[width=\textwidth]{figures/kitchen_kitchen-mixed-v0_TaskCompletionRate_zoom.pdf}
    
%     {\small (b) Mixed}
% \end{minipage}
% \hfill
% \begin{minipage}{0.30\textwidth}
%     \centering
%     \includegraphics[width=\textwidth]{figures/kitchen_kitchen-partial-v0_TaskCompletionRate_no_zoom.pdf}
    
%     {\small (c) Partial}
% \end{minipage}

% \vspace{0.2em}
% \includegraphics[width=0.6\textwidth]{figures/kitchen_kitchen-complete-v0_SuccessRate_legend-v2.pdf}

% \vspace{-0.8em}
% \captionof{figure}{Learning curves on Franka Kitchen multi-task benchmark.}
% \label{fig:kitchen}
% \vspace{1em}
% }]

\subsection{Results}
\label{sec:experiments_results}

% Table~\ref{tab:main_results} summarizes the quantitative results in all benchmarks. Figures~\ref{fig:d4rl}, \ref{fig:robomimic}, and \ref{fig:kitchen} show the corresponding learning curves. detailed configurations and the rationale for these choices are provided in Appendix~\ref{app:exp_details} and Table~\ref{tab:model_variants} 

Table~\ref{tab:main_results} summarizes the quantitative results in all benchmarks. Figures~\ref{fig:d4rl}, \ref{fig:robomimic}, and \ref{fig:kitchen} show the corresponding learning curves. Detailed configurations and rationale for the selection of the backbone model, based on the settings and training performance reported in ReinFlow~\citep{zhang2025reinflow}, are provided in Table~\ref{tab:model_variants} in Appendix~\ref{app:exp_details}.

\paragraph{D4RL Locomotion.}

ScoRe-Flow achieves the highest average performance across all D4RL tasks, outperforming both the ReinFlow variants and DPPO. 
In particular, ScoRe-Flow excels in the challenging Humanoid-v3 task with 5100$\pm$47 reward, surpassing all baselines including the strong ReinFlow-R (5041) and DPPO (5001). 
In terms of efficiency, ScoRe-Flow achieves an average time-to-convergence of just 0.16 hours to reach 90\% of the final performance, compared to DPPO's 3.48-hour average. We note that this wall-clock speedup is primarily a \emph{structural} advantage shared by all flow-based methods over diffusion-based methods, as flow policies require only 2--4 denoising steps versus the 50--100 steps typical of DPPO. More relevant to our algorithmic contribution, ScoRe-Flow achieves $2.4\times$ faster convergence than the flow-based SOTA (ReinFlow) at the same $K=4$ denoising steps. We attribute this to the score-based drift modulation, which provides per-step geometric guidance toward high-density regions, accelerating early-stage learning before $\mathbf{v}_\theta$ is well-calibrated.

\paragraph{Robomimic.}

ScoRe-Flow consistently outperforms all baselines in PickPlaceCan, Square, and Transport tasks, achieving the highest success rates while maintaining competitive training efficiency. Despite challenges posed by visual observations, our score-based mean control effectively guides exploration in high-dimensional image spaces.

\paragraph{Franka Kitchen.}

Despite the challenging multi-task setting with sparse rewards, ScoRe-Flow achieves substantial improvements over baselines, demonstrating the effectiveness of our unified mean-variance control. The Score-SDE ablation achieves faster convergence but lower final performance, confirming that combining score-based mean control with learned variance prediction yields the optimal balance.

\paragraph{Evaluation Protocol.}
% We report learning curves showing the episode reward for D4RL tasks and the success rate for Robomimic experiments versus environment steps.
% For Franka Kitchen, we report the average tasks completed (out of 4) in Table~\ref{tab:main_results}, while Figure~\ref{fig:kitchen} plots the \textbf{success rate} normalized to $[0, 1]$ for visualization.
% All experiments are carried out with 3 random seeds, and we report the mean standard deviation $\pm$.
 
Table~\ref{tab:main_results} summarizes quantitative performance, reporting episode rewards for D4RL locomotion and success rates for Robomimic manipulation.
For Franka Kitchen, the table details average subtasks completed out of 4 to highlight granularity, while Figure~\ref{fig:kitchen} plots the success rate normalized to $[0, 1]$ for visualizing convergence.
All methods use identical pre-trained policy initialization and computational budget for fair comparison.

% D4RL Time data backup (hours to reach 90% performance):
% Hopper-v2:    DPPO=11.28, ReinFlow-R=1.10, ReinFlow-S=0.60, Score-SDE=0.12, ScoRe-Flow=0.28
% Walker2d-v2:  DPPO=0.66,  ReinFlow-R=0.15, ReinFlow-S=0.11, Score-SDE=0.04, ScoRe-Flow=0.10
% Ant-v0:       DPPO=0.66,  ReinFlow-R=0.32, ReinFlow-S=0.39, Score-SDE=0.10, ScoRe-Flow=0.02
% Humanoid-v3:  DPPO=1.32,  ReinFlow-R=0.40, ReinFlow-S=0.42, Score-SDE=0.49, ScoRe-Flow=0.23

\section{Analysis}

\begin{figure}[!t]
    \centering
    \includegraphics[width=0.9\columnwidth]{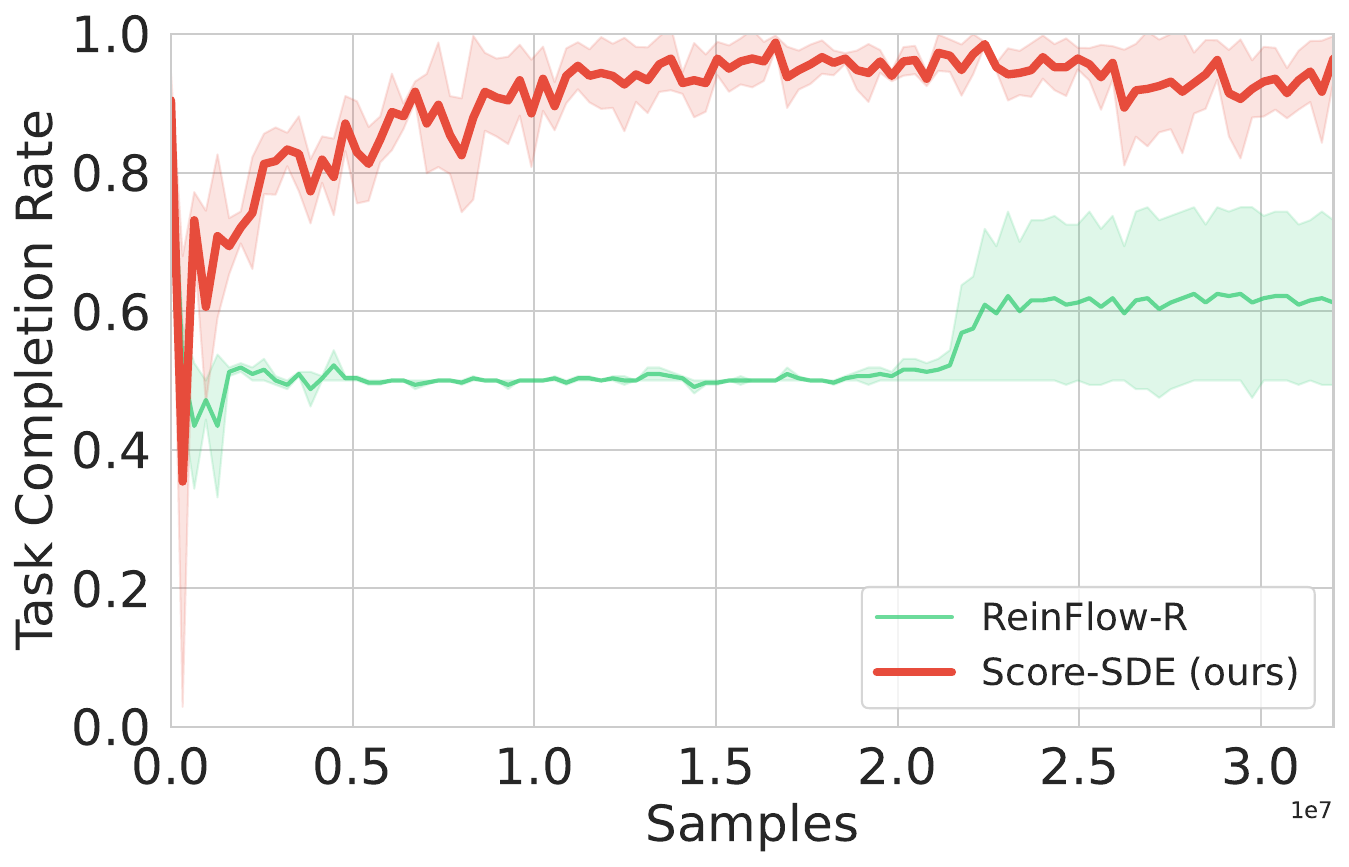}
    \captionof{figure}{Training stability comparison on Kitchen-Complete-v0. Under the same Rectified Flow base policy, ScoRe-Flow achieves stable learning while ReinFlow exhibits high variance and unstable convergence behavior.}
    \label{fig:score_comparison}
\end{figure}

\paragraph{Training Stability Comparison.}
A key advantage of ScoRe-Flow is improved training stability. Figure~\ref{fig:score_comparison} compares learning dynamics on Kitchen-Complete using Rectified Flow. While ReinFlow exhibits unstable learning with high variance, ScoRe-Flow maintains stable improvement throughout training. This stability stems from score-based mean control, which provides geometrically grounded guidance preventing the policy from drifting into low-probability regions.

% \paragraph{Sensitivity of Score-Only Methods to Noise Variance.}

% While score-based mean control provides effective exploration guidance, using it alone (Score-SDE) without learned variance prediction introduces a critical limitation: the method becomes highly sensitive to the manually specified noise variance $\sigma^2$. Figure~\ref{fig:std_ablation} shows the performance of Score-SDE under different fixed noise standard deviations on the Kitchen-Complete task. The optimal $\sigma$ value varies significantly across tasks and must be carefully tuned for each environment. Too small $\sigma$ leads to insufficient exploration, while too large $\sigma$ causes instability and performance degradation. In contrast, ScoRe-Flow learns the variance adaptively through $\sigma_\phi$, eliminating the need for task-specific noise tuning and achieving robust performance across diverse environments.

% \vspace{0.5em}
% \begin{center}
%     \includegraphics[width=0.9\columnwidth]{figures/score-based-SDE_method_std_ablation.pdf}
%     \captionof{figure}{Sensitivity of Score-SDE to initial noise variance. Performance varies dramatically with different fixed $\sigma$ values, demonstrating that score-only methods require careful, task-specific noise tuning.}
%     \label{fig:std_ablation}
% \end{center}

\paragraph{Sensitivity of Score-Only Methods to Noise Variance.}

\begin{figure}[!t]
    \centering
    \includegraphics[width=\columnwidth]{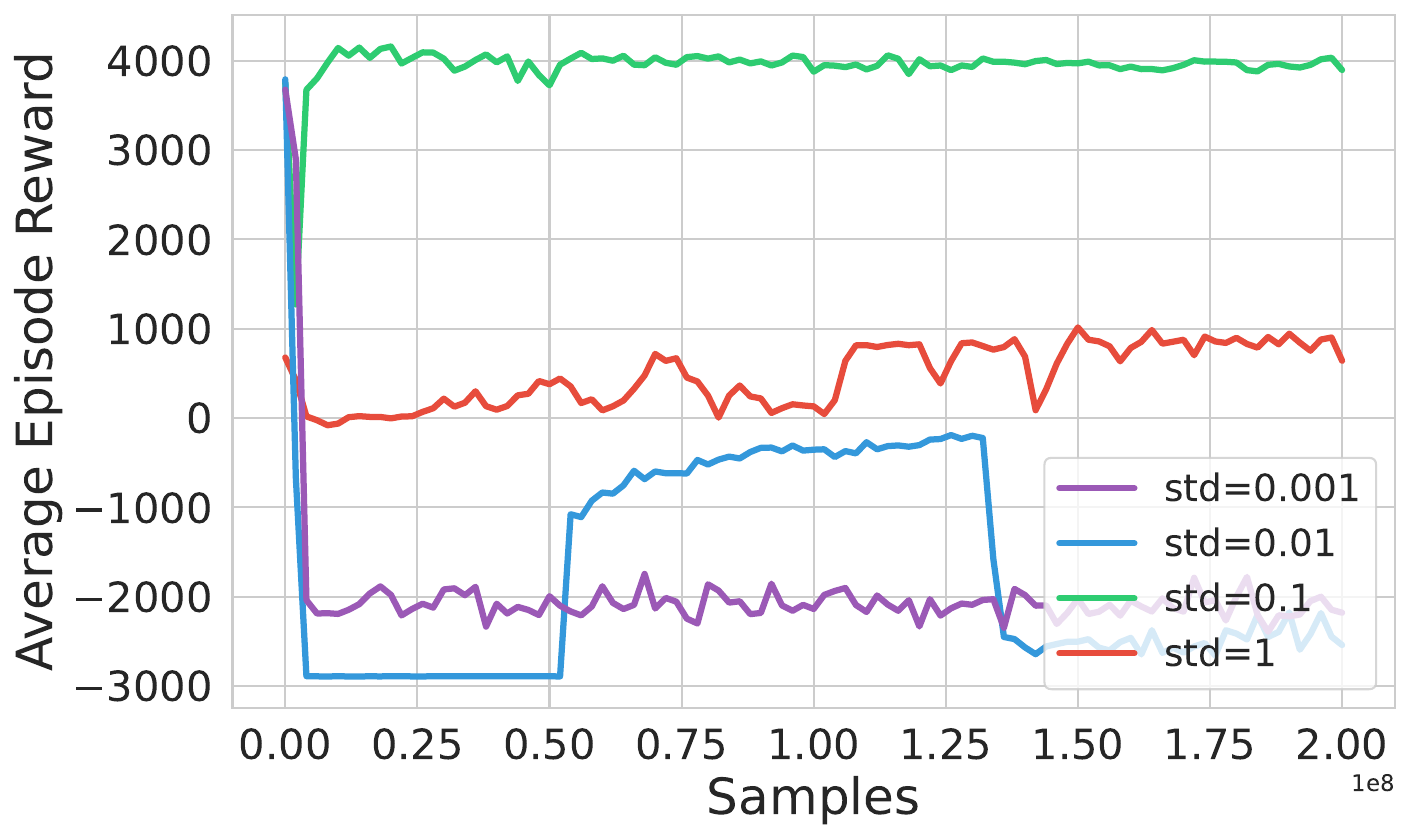}
    \captionof{figure}{Sensitivity of Score-SDE to initial noise variance. Performance varies dramatically with different fixed $\sigma$ values, demonstrating that score-only methods require careful task-specific tuning.}
    \label{fig:std_ablation}
\end{figure}

Using Score-SDE alone without learned variance prediction makes the method highly sensitive to manually specified noise variance $\sigma^2$. Figure~\ref{fig:std_ablation} shows Score-SDE performance under different fixed $\sigma$ on Ant-v0. The optimal $\sigma$ varies significantly between tasks: too small leads to insufficient exploration, while too large causes instability. ScoRe-Flow learns variance adaptively through $\sigma_\phi$, eliminating task-specific tuning.

\paragraph{Additional Ablation Studies.}
We provide comprehensive ablation studies in Appendix~D, analyzing the impact of denoising steps ($K$) and the decoupled variance prediction mechanism. Results show that while Score-SDE's rigid linear decay offers fast initial convergence, it lacks adaptability and yields lower asymptotic performance than ScoRe-Flow's learned variance.

\section{Conclusion}

We presented ScoRe-Flow, a method for RL fine-tuning of Flow Matching policies achieving decoupled mean-variance control. Building on the known score-velocity duality for linear flow paths \citep{albergo2023stochastic, lipman2022flow}, we enable score-based drift modulation with negligible overhead. Combined with learned variance prediction, ScoRe-Flow independently adjusts exploration direction and magnitude. Experiments demonstrate faster convergence and improved performance over variance-only and score-only baselines.

\paragraph{Limitations.}
The closed-form score expression (Eq.~\ref{eq:score_closed_form}) relies on the Gaussian conditional structure of linear interpolation paths with a Gaussian source distribution. For non-linear flow paths (e.g., Riemannian Flow Matching) or non-Gaussian source distributions, the conditional density is no longer Gaussian, and the score would require approximate or learned estimation — potentially losing the zero-auxiliary-network advantage. Linear paths cover the dominant use case in current robotic control applications \citep{lipman2022flow, zhang2025reinflow}, but extending to general stochastic interpolants \citep{albergo2023stochastic} is an important direction for future work.

\section*{Impact Statement}

This paper presents work whose goal is to advance the field of Machine
Learning. There are many potential societal consequences of our work, none
which we feel must be specifically highlighted here.

% \clearpage
\bibliography{reference}
\bibliographystyle{icml2026}

%%%%%%%%%%%%%%%%%%%%%%%%%%%%%%%%%%%%%%%%%%%%%%%%%%%%%%%%%%%%%%%%%%%%%%%%%%%%%%%
%%%%%%%%%%%%%%%%%%%%%%%%%%%%%%%%%%%%%%%%%%%%%%%%%%%%%%%%%%%%%%%%%%%%%%%%%%%%%%%
% APPENDIX
%%%%%%%%%%%%%%%%%%%%%%%%%%%%%%%%%%%%%%%%%%%%%%%%%%%%%%%%%%%%%%%%%%%%%%%%%%%%%%%
%%%%%%%%%%%%%%%%%%%%%%%%%%%%%%%%%%%%%%%%%%%%%%%%%%%%%%%%%%%%%%%%%%%%%%%%%%%%%%%
\newpage
\appendix
\onecolumn

\section{Derivation of the Marginal Score for Linear Flow Matching}
\label{app:score_derivation}

We derive a closed-form expression for the \emph{marginal} score
$\mathbf{s}_t(\mathbf{a}) := \nabla_{\mathbf{a}} \log \rho_t(\mathbf{a})$
along a linear Flow Matching (FM) probability path, conditioned on an observation $\mathbf{s}$
(omitted when clear).

\paragraph{Setup.}
Let $\mathbf{a}_0 \sim \mathcal{N}(\mathbf{0}, \mathbf{I})$ and $\mathbf{a}_1 \sim p_{\text{data}}(\cdot \mid \mathbf{s})$.
For $t \in [0,1)$, define the linear interpolation
\begin{equation}
\mathbf{a}_t = (1-t)\mathbf{a}_0 + t\mathbf{a}_1 .
\label{eq:linear_path}
\end{equation}
Define the per-sample (conditional) target velocity
\begin{equation}
\mathbf{v}^\star := \mathbf{a}_1 - \mathbf{a}_0 ,
\qquad\text{so that}\qquad
\mathbf{a}_t = \mathbf{a}_0 + t\mathbf{v}^\star .
\label{eq:vstar_def}
\end{equation}
Let $\rho_t(\mathbf{a})$ denote the \emph{marginal} density of $\mathbf{a}_t$ under the joint sampling
$\mathbf{a}_0 \sim \mathcal{N}(\mathbf{0},\mathbf{I})$, $\mathbf{a}_1 \sim p_{\text{data}}(\cdot\mid \mathbf{s})$.

\paragraph{Step 1: Conditional density and conditional score.}
Conditioned on $\mathbf{a}_1$, \eqref{eq:linear_path} becomes an affine transformation of Gaussian noise:
\begin{equation}
\mathbf{a}_t \mid \mathbf{a}_1 \sim \mathcal{N}\big(t\mathbf{a}_1,\; (1-t)^2 \mathbf{I}\big).
\label{eq:conditional_density}
\end{equation}
Hence the conditional log-density is
\begin{equation}
\log \rho_t(\mathbf{a}_t \mid \mathbf{a}_1)
= -\frac{1}{2(1-t)^2}\left\|\mathbf{a}_t - t\mathbf{a}_1\right\|^2 + C(t),
\end{equation}
and the conditional score is
\begin{equation}
\nabla_{\mathbf{a}_t}\log \rho_t(\mathbf{a}_t \mid \mathbf{a}_1)
= -\frac{\mathbf{a}_t - t\mathbf{a}_1}{(1-t)^2}.
\label{eq:conditional_score}
\end{equation}

\paragraph{Step 2: Marginal score as a posterior expectation (no marginal/conditional conflation).}
The marginal density is $\rho_t(\mathbf{a}_t) = \int \rho_t(\mathbf{a}_t\mid \mathbf{a}_1)p_{\text{data}}(\mathbf{a}_1\mid \mathbf{s})\,d\mathbf{a}_1$.
Differentiating under the integral sign yields the standard identity:
\begin{align}
\nabla_{\mathbf{a}_t} \log \rho_t(\mathbf{a}_t)
&= \frac{1}{\rho_t(\mathbf{a}_t)} \nabla_{\mathbf{a}_t} \int \rho_t(\mathbf{a}_t\mid \mathbf{a}_1)p_{\text{data}}(\mathbf{a}_1\mid \mathbf{s})\,d\mathbf{a}_1 \nonumber \\
&= \int \frac{\rho_t(\mathbf{a}_t\mid \mathbf{a}_1)p_{\text{data}}(\mathbf{a}_1\mid \mathbf{s})}{\rho_t(\mathbf{a}_t)}
\;\nabla_{\mathbf{a}_t}\log \rho_t(\mathbf{a}_t\mid \mathbf{a}_1)\; d\mathbf{a}_1 \nonumber \\
&= \mathbb{E}\!\left[\,\nabla_{\mathbf{a}_t}\log \rho_t(\mathbf{a}_t\mid \mathbf{a}_1)\ \big|\ \mathbf{a}_t,\mathbf{s}\right].
\label{eq:marginal_score_identity}
\end{align}
Substituting \eqref{eq:conditional_score} into \eqref{eq:marginal_score_identity} gives an exact expression:
\begin{equation}
\mathbf{s}_t(\mathbf{a}_t)
= -\frac{\mathbf{a}_t - t\,\mathbb{E}[\mathbf{a}_1 \mid \mathbf{a}_t,\mathbf{s}]}{(1-t)^2}.
\label{eq:marginal_score_via_a1}
\end{equation}

\paragraph{Step 3: Rewriting in terms of the (marginal) velocity field.}
Using the path relation $\mathbf{a}_t=(1-t)\mathbf{a}_0+t\mathbf{a}_1$, taking conditional expectation given $(\mathbf{a}_t,\mathbf{s})$ yields
\begin{equation}
\mathbb{E}[\mathbf{a}_0 \mid \mathbf{a}_t,\mathbf{s}]
= \frac{\mathbf{a}_t - t\,\mathbb{E}[\mathbf{a}_1 \mid \mathbf{a}_t,\mathbf{s}]}{1-t}.
\label{eq:posterior_a0}
\end{equation}
Therefore the posterior mean of the target velocity $\mathbf{v}^\star=\mathbf{a}_1-\mathbf{a}_0$ is
\begin{align}
\mathbb{E}[\mathbf{v}^\star \mid \mathbf{a}_t,\mathbf{s}]
&= \mathbb{E}[\mathbf{a}_1 \mid \mathbf{a}_t,\mathbf{s}] - \mathbb{E}[\mathbf{a}_0 \mid \mathbf{a}_t,\mathbf{s}] \nonumber \\
&= \frac{\mathbb{E}[\mathbf{a}_1 \mid \mathbf{a}_t,\mathbf{s}] - \mathbf{a}_t}{1-t}.
\label{eq:posterior_vstar}
\end{align}
Plugging \eqref{eq:posterior_vstar} into \eqref{eq:marginal_score_via_a1} yields an equivalent and often more convenient form:
\begin{equation}
\boxed{
\mathbf{s}_t(\mathbf{a}_t)
= \frac{t\,\mathbb{E}[\mathbf{v}^\star \mid \mathbf{a}_t,\mathbf{s}] - \mathbf{a}_t}{1-t}.
}
\label{eq:marginal_score_via_vstar}
\end{equation}

\paragraph{Step 4: Practical estimator via a learned FM velocity field.}
Flow Matching learns a \emph{marginal} vector field, which (under the standard conditional-to-marginal construction)
corresponds to the posterior average of per-sample conditional velocities. Concretely, for the linear path,
the optimal marginal velocity satisfies
\begin{equation}
\mathbf{v}_{\text{marg}}(t,\mathbf{a}_t,\mathbf{s})
= \mathbb{E}[\mathbf{v}^\star \mid \mathbf{a}_t,\mathbf{s}].
\label{eq:marginal_velocity_equals_posterior_mean}
\end{equation}
Thus, with a learned FM velocity field $\mathbf{v}_\theta(t,\mathbf{a}_t,\mathbf{s}) \approx \mathbf{v}_{\text{marg}}(t,\mathbf{a}_t,\mathbf{s})$,
we obtain the score estimator used in the main text:
\begin{equation}
\boxed{
\mathbf{s}_t(\mathbf{a}_t)
\approx \frac{t\,\mathbf{v}_\theta(t,\mathbf{a}_t,\mathbf{s}) - \mathbf{a}_t}{1-t}.
}
\label{eq:score_estimator}
\end{equation}

% \paragraph{Asymptotic behavior near $t\to 1$ and stabilization.}
% The prefactor $(1-t)^{-1}$ in \eqref{eq:score_estimator} implies $\|\mathbf{s}_t(\mathbf{a}_t)\| = O((1-t)^{-1})$ as $t\to 1$
% (for bounded $\mathbf{v}_\theta$). This motivates annealing the coefficient of the score correction (e.g., scaling it as $(1-t)^\gamma$ with $\gamma\ge 1$)
% so that the score-modulated drift remains bounded and the induced stochastic dynamics are numerically stable.
\paragraph{Asymptotic behavior near $t\to 1$ and stabilization.}
The prefactor $(1-t)^{-1}$ in \eqref{eq:score_estimator} implies $\|\mathbf{s}_t(\mathbf{a}_t)\| = O((1-t)^{-1})$ as $t\to 1$ (for bounded $\mathbf{v}_\theta$). 
Therefore, the coefficient multiplying the score in the drift should decay as $O(1-t)$ to keep the score-modulated drift bounded. In ScoRe-Flow, we implement this stabilization by enforcing a hard time-decay in the score scheduler (see \cref{eq:alpha_scaled}), i.e., $\alpha_\psi^{\text{scaled}}(t)\propto (1-t)$. For score-only baselines where the same schedule also determines the diffusion (Score-SDE), we use linear decay as a canonical instance.

\section{Network Architecture Details}
\label{app:architecture}

\paragraph{Score Scheduler ($\alpha_\psi$).}
The score scheduler takes as input only the flow time $t \in [0, 1]$ and outputs a scalar weight $\alpha_\psi(t) > 0$. The architecture consists of a 2-layer MLP and SiLU activations, followed by a Softplus output activation to ensure positivity. A hard time-decay constraint is applied:
\begin{equation}
\alpha_\psi^{\text{scaled}}(t) = (1-t) \cdot \alpha_\psi(t).
\end{equation}
This ensures $\alpha_\psi^{\text{scaled}} \to 0$ as $t \to 1$, preventing numerical instability from the score function's $(1-t)^{-1}$ term

\paragraph{Variance Predictor ($\sigma_\phi$).}
The variance predictor takes as input the condition embedding $\mathbf{c} \in \mathbb{R}^{d_c}$ from the policy encoder (which encodes observation history). The architecture consists of an MLP with hidden dimensions specified per task (typically $[64, 64]$ for state-based tasks and $[256, 256, 256]$ for vision tasks), using Tanh activations. The output is mapped to a bounded range $[\sigma_{\min}, \sigma_{\max}]$ via:
\begin{equation}
\sigma_\phi = \sigma_{\min} + \frac{\sigma_{\max} - \sigma_{\min}}{2} \cdot (\tanh(\text{MLP}(\mathbf{c})) + 1)
\end{equation}

\paragraph{Base Velocity Field ($\mathbf{v}_\theta$).}
We use the same architecture as the pre-trained FM policy, typically a transformer-based architecture for vision tasks or an MLP for state-based tasks. During RL fine-tuning, $\mathbf{v}_\theta$ is jointly optimized with $(\alpha_\psi, \sigma_\phi)$.

\paragraph{Initialization.}
The score scheduler's final linear layer is initialized with zero weights and bias $-2.0$, so that $\text{Softplus}(-2) \approx 0.13$, ensuring small initial $\alpha$ values for training stability.

\section{Hyperparameters}
\label{app:hyperparams}
In this section, we provide a comprehensive listing of the hyperparameters used for training \textbf{ScoRe-Flow} and the \textbf{Score-based SDE} baseline. To ensure reproducibility and facilitate a rigorous comparison, we align our base configurations, such as network architectures, optimizer settings, and learning rate schedules, with established baselines (e.g., ReinFlow, DPPO) wherever applicable. The configurations are organized hierarchically: \cref{tab:shared_params_detailed} outlines the shared parameters common across all tasks, while \cref{tab:gym_detailed_params}, \cref{tab:kitchen_params_combined}, and \cref{tab:robomimic_params_combined} detail the specific settings tailored to the unique dynamics of the OpenAI Gym, Franka Kitchen, and Robomimic benchmarks, respectively.
% ==========================================
% Table 6: Shared Hyperparameters Across All Tasks
% ==========================================
\begin{table}[hbt!]
    \centering
    \caption{Shared Hyperparameter Configurations for Score-based SDE and ScoRe-Flow.}
    \label{tab:shared_params_detailed}
    \small
    \renewcommand{\arraystretch}{1.1}
    \begin{tabular}{lc}
        \toprule
        \textbf{Parameter} & \textbf{Value (Common to Both)} \\
        \midrule
        \multicolumn{2}{l}{\textit{Optimizer \& Learning Rate Settings}} \\
        Actor Optimizer & Adam\citep{kingma2014adam} \\
        Actor LR Weight Decay & 0 \\
        Actor LR Scheduler & CosineAnnealingWarmupRestart\citep{loshchilov2016sgdr} \\
        Actor LR Cycle Steps & 100$^\ast$ \\
        \addlinespace
        Critic Optimizer & Adam \\
        Critic LR Scheduler & CosineAnnealingWarmupRestart \\
        Critic Scheduler Warmup & 10 \\
        Critic LR Cycle Steps & 100$^\dagger$ \\
        \midrule
        \multicolumn{2}{l}{\textit{Network Architecture \& Training}} \\
        Policy Network Architecture & 512 $\times$ 3 MLP \\
        Critic Network Architecture & 256 $\times$ 3 MLP \\
        Critic Loss Coef. & 0.50 \\
        Reward Scale & 1.0 \\
        Reward Normalization & True \\
        \bottomrule
        \multicolumn{2}{l}{\footnotesize $^\ast$Set to 150 for the Franka Kitchen-Partial task.} \\
        \multicolumn{2}{l}{\footnotesize $^\dagger$Set to 50 for the Franka Kitchen-Partial task.}
    \end{tabular}
\end{table}

% ==========================================
% Table 7: OpenAI Gym Tasks
% ==========================================
\begin{table}[hbt!]
    \centering
    \caption{Detailed Hyperparameter Settings for OpenAI Gym Locomotion Tasks.}
    \label{tab:gym_detailed_params}
    \small
    \renewcommand{\arraystretch}{1.0}

    % --- (a) Shared Gym Params ---
    \subcaptionbox{Shared and Algorithm-Specific Base Hyperparameters}{
        \centering
        \begin{tabular}{lcc}
            \toprule
            \textbf{Parameter} & \textbf{Score-based SDE} & \textbf{ScoRe-Flow} \\
            \midrule
            \multicolumn{3}{l}{\textit{PPO Optimization Configurations}} \\
            % --- 明确使用 epsilon_clip ---
            PPO Clip Ratio ($\epsilon_{clip}$) & \multicolumn{2}{c}{0.01} \\
            Target KL Divergence & \multicolumn{2}{c}{1.0} \\
            Entropy Coef. ($\alpha$) & \multicolumn{2}{c}{0.03} \\
            Num. Parallel Envs & \multicolumn{2}{c}{40} \\
            Batch Size & \multicolumn{2}{c}{50,000} \\
            Update Epochs & \multicolumn{2}{c}{5} \\
            Discount Factor ($\gamma$) & \multicolumn{2}{c}{0.99} \\
            GAE Lambda\citep{schulman2015high} ($\lambda$) & \multicolumn{2}{c}{0.95} \\
            Action Chunking Size\citep{lai2022action} & \multicolumn{2}{c}{4} \\
            Condition Stacking & \multicolumn{2}{c}{1} \\
            Denoising Steps ($K$) & \multicolumn{2}{c}{4} \\
            BC Loss Coef ($\beta$) & \multicolumn{2}{c}{0.01} \\
            Clip Interm. Actions & \multicolumn{2}{c}{True} \\
            Max Ep. Steps / Rollout Steps & \multicolumn{2}{c}{1000 / 500} \\
            \midrule
            \multicolumn{3}{l}{\textit{Exploration Mechanism}} \\
            Exploration Noise Type & $\epsilon$-SDE Stochastic & Learnable Noise Scheduler \\
            % --- 明确 Exploration epsilon schedule ---
            Exploration $\epsilon_t$ Schedule & Linear Decay & N/A \\
            Noise Hold Ratio & N/A & 35\% of total iteration \\
            Noise Decay Target & N/A & $0.3\sigma_{\min} + 0.7\sigma_{\max}$ \\
            \midrule
            \multicolumn{3}{l}{\textit{Score Parameters}} \\
            Learnable Score Scheduler & N/A & MLP \\
            Score Scheduler Hidden Dim & N/A & 16 \\
            \bottomrule
        \end{tabular}
    }

    \vspace{4mm}

    % --- (b) ScoRe-Flow Task Specific ---
    \subcaptionbox{Task-Specific Hyperparameters: ScoRe-Flow}{
        \centering
        \begin{tabular}{lcccc}
            \toprule
            \textbf{Parameter} & \textbf{Hopper-v2} & \textbf{Walker2d-v2} & \textbf{Ant-v0} & \textbf{Humanoid-v3} \\
            \midrule
            Min / Max Noise Std ($\sigma$) & 0.10 / 0.24 & 0.10 / 0.24 & 0.08 / 0.16 & 0.08 / 0.16 \\
            Critic Warmup Iters & 0 & 5 & 0 & 0 \\
            Actor LR (Cosine) & cos(4.5e-5, 2e-5) & cos(4e-4, 4e-4) & cos(4.5e-5, 2e-5) & cos(4.5e-5, 2e-5) \\
            Critic LR (Cosine) & cos(6.5e-4, 3e-4) & cos(4e-3, 4e-3) & cos(6.5e-4, 3e-4) & cos(6.5e-4, 3e-4) \\
            Total Training Iters & 1001 & 1001 & 1001 & 201 \\
            \bottomrule
        \end{tabular}
    }

    \vspace{4mm}

    % --- (c) Score-based SDE Task Specific ---
    \subcaptionbox{Task-Specific Hyperparameters: Score-based SDE}{
        \centering
        \begin{tabular}{lcccc}
            \toprule
            \textbf{Parameter} & \textbf{Hopper-v2} & \textbf{Walker2d-v2} & \textbf{Ant-v0} & \textbf{Humanoid-v3} \\
            \midrule
            % --- 明确使用 epsilon_t ---
            Base Exploration Coef. ($\epsilon_0$) & 0.1 & 0.1 & 0.01 & 0.01 \\
            Critic Warmup Iters & 0 & 3 to 5 & 3 to 5 & 3 to 5 \\
            Actor LR (Cosine) & cos(4.5e-5, 2e-5) & cos(4e-4, 4e-4) & cos(4e-4, 2e-4) & cos(4.5e-5, 2e-5) \\
            Critic LR (Cosine) & cos(6.5e-4, 3e-4) & cos(4e-3, 4e-3) & cos(4e-3, 2e-3) & cos(6.5e-4, 3e-4) \\
            Total Training Iters & 1001 & 1001 & 1001 & 201 \\
            \bottomrule
        \end{tabular}
    }
\end{table}

\begin{table}[hbt!]
    \centering
    \caption{Hyperparameter Configurations for Franka Kitchen Tasks.}
    \label{tab:kitchen_params_combined}
    \small
    \renewcommand{\arraystretch}{1.0}

    % --- (a) Shared Kitchen Params ---
    \subcaptionbox{Shared and Algorithm-Specific Base Hyperparameters}{
        \centering
        \begin{tabular}{lcc}
            \toprule
            \textbf{Parameter} & \textbf{Score-based SDE} & \textbf{ScoRe-Flow} \\
            \midrule
            \multicolumn{3}{l}{\textit{PPO Optimization Configurations}} \\
            % --- 明确使用 epsilon_clip ---
            PPO Clip Ratio ($\epsilon_{clip}$) & \multicolumn{2}{c}{0.01} \\
            Target KL Divergence & \multicolumn{2}{c}{1.0} \\
            Update Epochs & \multicolumn{2}{c}{10} \\
            Discount Factor ($\gamma$) & \multicolumn{2}{c}{0.99} \\
            GAE Lambda ($\lambda$) & \multicolumn{2}{c}{0.95} \\
            Num. Parallel Envs & \multicolumn{2}{c}{40} \\
            Batch Size & \multicolumn{2}{c}{5600} \\
            Total Training Iters & \multicolumn{2}{c}{301} \\
            Action Chunking Size & \multicolumn{2}{c}{4} \\
            Condition Stacking & \multicolumn{2}{c}{1} \\
            Denoising Steps ($K$) & \multicolumn{2}{c}{4} \\
            Clip Interm. Actions & \multicolumn{2}{c}{True} \\
            Random / Denoised Clip Value & \multicolumn{2}{c}{3.0 / 1.0} \\
            BC Loss Coef ($\beta$) & \multicolumn{2}{c}{0.00} \\
            Critic Weight Decay & \multicolumn{2}{c}{$1 \times 10^{-5}$} \\
            Critic Warmup Iters & \multicolumn{2}{c}{0} \\
            \midrule
            \multicolumn{3}{l}{\textit{Exploration Mechanism}} \\
            Exploration Noise Type & $\epsilon$-SDE Stochastic & Learnable Noise Scheduler \\
            % --- 明确 Exploration epsilon schedule ---
            Exploration $\epsilon_t$ Schedule & Linear Decay & N/A \\
            Noise Hold Ratio & N/A & \textbf{100\% of total iteration} \\
            Noise Decay Target & N/A & $\sigma_{\max}$ \\
            \midrule
            \multicolumn{3}{l}{\textit{Score Parameters}} \\
            Learnable Score Scheduler & N/A & MLP \\
            Score Scheduler Hidden Dim & N/A & 16 \\
            \bottomrule
        \end{tabular}
    }

    \vspace{4mm}

% --- (b) ScoRe-Flow Kitchen Task Specific ---
    \subcaptionbox{Task-Specific Hyperparameters: ScoRe-Flow}{
        \centering
        \begin{tabular}{lccc}
            \toprule
            \textbf{Parameter} & \textbf{Complete} & \textbf{Mixed} & \textbf{Partial} \\
            \midrule
            Finetuning Denoising Steps & 4 & 4 & 8 \\
            Min / Max Noise Std ($\sigma$) & 0.05 / 0.12 & 0.05 / 0.12 & 0.05 / 0.35 \\
            Entropy Coef ($\alpha$) & 0.00 & 0.00 & 0.01 \\
            Actor LR (Cosine) & cos(4.5e-5, 2e-5) & cos(4.5e-5, 2e-5) & cos(4.5e-5, 2e-5) \\
            Critic LR (Cosine) & cos(6.5e-4, 3e-4) & cos(6.5e-4, 3e-4) & cos(6.5e-4, 3e-4) \\
            \bottomrule
        \end{tabular}
    }

    \vspace{4mm}

    % --- (c) Score-based SDE Kitchen Task Specific ---
    \subcaptionbox{Task-Specific Hyperparameters: Score-based SDE}{
        \centering
        \begin{tabular}{lccc}
            \toprule
            \textbf{Parameter} & \textbf{Complete} & \textbf{Mixed} & \textbf{Partial} \\
            \midrule
            Base Exploration Coef. ($\epsilon_0$) & 0.01 to 0.1 & 0.01 & 0.05 \\
            Entropy Coef ($\alpha$) & 0.00 & 0.00 & 0.00 \\
            Actor LR (Cosine) & cos(4.5e-5, 2e-5) & cos(4.5e-5, 2e-5) & cos(4.5e-5, 2e-5) \\
            Critic LR (Cosine) & cos(6.5e-4, 3e-4) & cos(6.5e-4, 3e-4) & cos(2e-4, 5e-5) \\
            \bottomrule
        \end{tabular}
    }
\end{table}

% ==========================================
% Table 9: Robomimic Visual Tasks (Standardized)
% ==========================================
\begin{table}[hbt!]
    \centering
    \caption{Hyperparameter Configurations for Robomimic Image-input Manipulation Tasks.}
    \label{tab:robomimic_params_combined}
    \small
    \renewcommand{\arraystretch}{1.0}

    % --- (a) Shared Robomimic Params ---
    \subcaptionbox{Shared and Algorithm-Specific Base Hyperparameters}{
        \centering
        \begin{tabular}{lcc}
            \toprule
            \textbf{Parameter} & \textbf{Score-based SDE} & \textbf{ScoRe-Flow} \\
            \midrule
            \multicolumn{3}{l}{\textit{PPO Optimization Configurations}} \\
            PPO Clip Ratio ($\epsilon_{clip}$) & \multicolumn{2}{c}{0.01} \\
            % Robomimic 特有的 Target KL (0.01 vs 1.0)，必须保留
            Target KL Divergence & \multicolumn{2}{c}{0.01} \\
            % Robomimic 特有的 Update Epochs (10 vs 5)，必须保留
            Update Epochs & \multicolumn{2}{c}{10} \\
            % Robomimic 特有的 Gamma (0.999 vs 0.99)，必须保留
            Discount Factor ($\gamma$) & \multicolumn{2}{c}{0.999} \\
            GAE Lambda ($\lambda$) & \multicolumn{2}{c}{0.95} \\
            Num. Parallel Envs & \multicolumn{2}{c}{50} \\
            Batch Size & \multicolumn{2}{c}{500} \\
            Condition Stacking & \multicolumn{2}{c}{1} \\
            Clip Interm. Actions & \multicolumn{2}{c}{True} \\
            Random / Denoised Clip Value & \multicolumn{2}{c}{3.0 / 1.0} \\
            BC Loss Coef ($\beta$) & \multicolumn{2}{c}{0.00} \\
            Max Grad Norm & \multicolumn{2}{c}{25.0} \\
            \midrule
            \multicolumn{3}{l}{\textit{Exploration Mechanism}} \\
            Exploration Noise Type & $\epsilon$-SDE Stochastic & Learnable Noise Scheduler \\
            Exploration $\epsilon_t$ Schedule & Linear Decay & N/A \\
            Noise Scheduler Type & N/A & Learnable Decay \\
            Noise Hold Ratio & N/A & \textbf{100\% of total iteration} \\
            Noise Decay Target & N/A & $\sigma_{\max}$ \\
            \midrule
            \multicolumn{3}{l}{\textit{Score Parameters}} \\
            Learnable Score Scheduler & N/A & MLP \\
            Score Scheduler Hidden Dim & N/A & 16 \\
            \bottomrule
        \end{tabular}
    }

    \vspace{4mm}

% --- (b) ScoRe-Flow Robomimic Task Specific ---
    \subcaptionbox{Task-Specific Hyperparameters: ScoRe-Flow}{
        \centering
        \begin{tabular}{lccc}
            \toprule
            \textbf{Parameter} & \textbf{Can} & \textbf{Square} & \textbf{Transport} \\
            \midrule
            Total Training Iters & 151 & 301 & 201 \\
            Action Chunking Size & 4 & 4 & 8 \\
            Denoising Steps ($K$) & 1 & 4 & 4 \\
            Min / Max Noise Std ($\sigma$) & 0.08 / 0.14 & 0.08 / 0.14 & 0.05 / 0.10 \\
            Entropy Coef ($\alpha$) & 0.01 & 0.01 & 0.01 \\
            Critic Warmup Iters & 5 & 5 & 5 \\
            Actor LR (Cosine) & cos(2e-5, 1e-5) & cos(3.5e-6, 3.5e-6) & cos(3.5e-6, 3.5e-6) \\
            Critic LR (Cosine) & cos(6.5e-4, 3e-4) & cos(4.5e-4, 3e-4) & cos(3.2e-4, 3e-4) \\
            Critic Weight Decay & 0 & 0 & 1e-5 \\
            \bottomrule
        \end{tabular}
    }

    \vspace{4mm}

    % --- (c) Score-based SDE Robomimic Task Specific ---
    \subcaptionbox{Task-Specific Hyperparameters: Score-based SDE}{
        \centering
        \begin{tabular}{lccc}
            \toprule
            \textbf{Parameter} & \textbf{Can} & \textbf{Square} & \textbf{Transport} \\
            \midrule
            Total Training Iters & 151 & 301 & 201 \\
            Action Chunking Size & 4 & 4 & 8 \\
            Denoising Steps ($K$) & 1 & 2 & 4 \\
            Max Episode Steps & 300 & 400 & 800 \\
            Num. Rollout Steps & 300 & 400 & 400 \\
            Base Exploration Coef. ($\epsilon_0$) & 0.01& 0.01& 0.01 \\
            Entropy Coef ($\alpha$) & 0 & 0 & 0 \\
            Critic Warmup Iters & 2 & 2 & 5\\
            Actor LR (Cosine) & cos(2e-5, 1e-5) & cos(3.5e-6, 3.5e-6) & cos(3.5e-6, 3.5e-6) \\
            Critic LR (Cosine) & cos(6.5e-4, 3e-4) & cos(4.5e-4, 3e-4) & cos(3.2e-4, 3e-4) \\
            Critic Weight Decay & 0 & 0 & 1e-5 \\
            \bottomrule
        \end{tabular}
    }
\end{table}
% \todo{maybe Action Chunking Size is 2 in square ? need more experiment}

%------------------------------------------------------------------------------
\section{Additional Experimental Details}
\label{app:exp_details}

% \paragraph{Pre-training.}
% All methods start from the same pre-trained FM velocity field $\mathbf{v}_\theta$, trained via standard Flow Matching on demonstration datasets. For D4RL, we use the ``medium-expert'' dataset. For Robomimic and Franka Kitchen, we use the provided human demonstrations.

% \paragraph{Evaluation Protocol.}
% We evaluate policies every 10K environment steps using 10 deterministic rollouts (with $\sigma_\phi = 0$ and $\alpha_\psi = 0$ for fair comparison). We report the mean return with standard error over 5 random seeds.

% \paragraph{Compute Resources.}
% All experiments were conducted on a single NVIDIA A100 GPU. Training ScoRe-Flow for 1M steps takes approximately 8 hours on D4RL tasks and 12 hours on Robomimic tasks.

\paragraph{Random Seeds and Variability.}
We train all RL algorithms using 3 random seeds for all the tasks.  The RL fine-tuning curves report the average reward or success rate over these seeds, with shading representing the mean $\pm$ standard deviation to indicate variability between runs.

\paragraph{Rendering Backend.}
To accelerate GPU-based computation during training and wall-time testing, we set the MuJoCo graphics rendering backend (\texttt{MUJOCO\_GL}) to the Embedded System Graphics Library (EGL). We note that users relying on software rendering (e.g., \texttt{osmesa}) or running multiple threads on shared CPU/GPU kernels may experience longer compute times than those reported in our benchmarks.

\paragraph{Wall-clock Time Analysis.}
For fair comparison, wall-clock time is measured sequentially on a single NVIDIA RTX 3090 GPU with EGL rendering for all tasks, with the exception of Robomimic Transport where two A100 GPUs are employed. We calculate the total wall-clock time by multiplying the average iteration time by the total number of iterations. 

Quantitative results highlight the efficiency of our method: ScoRe-Flow reaches 90\% of final performance in only \textbf{0.16 hours on average} for D4RL tasks. This represents a \textbf{22$\times$ speedup} compared to DPPO (3.48 hours), which typically requires 50 to 100 denoising steps. This efficiency stems from our use of fewer denoising steps ($K=4$) and the effective guidance of score-based mean control. 

While FQL shows shorter per-iteration times, it is not necessarily more efficient overall: its batch size is considerably smaller than PPO-based methods, and it requires significantly more total training iterations to converge. Furthermore, FQL lacks the parallel computing design inherent to DPPO and ReinFlow.

\begin{table}[hbt!]
    \centering
    \caption{\textbf{Per Iteration Wall-clock Time Analysis on D4RL OpenAI Gym Locomotion Tasks (Seconds).} Comparison of single training iteration time across three random seeds. ScoRe-Flow maintains competitive computational efficiency compared to flow-based baselines, with only negligible overhead introduced by the variance prediction mechanism.}
    \label{tab:wall_clock_time}
    \footnotesize  % 保持小字体
    \setlength{\tabcolsep}{3.0pt} % 进一步微调列间距以容纳更长的名字
    \renewcommand{\arraystretch}{0.95}
    \begin{tabular}{llcccc}
        \toprule
        \multirow{2}{*}{\textbf{Task}} & \multirow{2}{*}{\textbf{Method}} & \multicolumn{3}{c}{\textbf{Single Iteration (s)}} & \multirow{2}{*}{\textbf{Mean $\pm$ Std}} \\
        \cmidrule(lr){3-5}
        & & \textbf{S1} & \textbf{S2} & \textbf{S3} & \\
        \midrule
        
        % Hopper (6 rows now)
        \multirow{6}{*}{Hopper-v2} 
        & ReinFlow-R & 11.598 & 11.704 & 11.843 & 11.715 $\pm$ 0.123 \\
        & ReinFlow-S & 12.051 & 12.127 & 12.372 & 12.290 $\pm$ 0.141 \\
        & DPPO       & 99.502 & 99.616 & 98.021 & 99.046 $\pm$ 0.890 \\
        & FQL        & 4.373  & 4.366  & 4.515  & 4.418 $\pm$ 0.084  \\
        & Score-SDE  & 11.652 & 11.715 & 11.793 & 11.720 $\pm$ 0.071 \\
        & \textbf{ScoRe-Flow} & \textbf{11.821} & \textbf{11.904} & \textbf{11.885} & \textbf{11.870 $\pm$ 0.043} \\
        \midrule
        
        % Walker (6 rows)
        \multirow{6}{*}{Walker2d-v2} 
        & ReinFlow-R & 11.861 & 11.446 & 11.382 & 11.563 $\pm$ 0.260 \\
        & ReinFlow-S & 12.393 & 12.690 & 13.975 & 13.019 $\pm$ 0.841 \\
        & DPPO       & 101.15 & 106.12 & 98.470 & 101.91 $\pm$ 3.884 \\
        & FQL        & 5.248  & 4.597  & 5.207  & 5.017 $\pm$ 0.365  \\
        & Score-SDE  & 11.605 & 11.524 & 11.588 & 11.572 $\pm$ 0.043 \\
        & \textbf{ScoRe-Flow} & \textbf{11.782} & \textbf{11.845} & \textbf{11.751} & \textbf{11.793 $\pm$ 0.048} \\
        \midrule
        
        % Ant (6 rows)
        \multirow{6}{*}{Ant-v0} 
        & ReinFlow-R & 17.210 & 17.685 & 17.524 & 17.473 $\pm$ 0.242 \\
        & ReinFlow-S & 17.291 & 17.821 & 18.090 & 17.734 $\pm$ 0.407 \\
        & DPPO       & 102.36 & 104.63 & 99.042 & 102.01 $\pm$ 2.811 \\
        & FQL        & 5.242  & 4.950  & 5.309  & 5.167 $\pm$ 0.191  \\
        & Score-SDE  & 17.355 & 17.422 & 17.501 & 17.426 $\pm$ 0.073 \\
        & \textbf{ScoRe-Flow} & \textbf{17.653} & \textbf{17.712} & \textbf{17.805} & \textbf{17.723 $\pm$ 0.077} \\
        \midrule
        
        % Humanoid (6 rows)
        \multirow{6}{*}{Humanoid-v3} 
        & ReinFlow-R & 31.437 & 30.223 & 31.088 & 30.916 $\pm$ 0.625 \\
        & ReinFlow-S & 30.499 & 30.058 & 31.029 & 30.529 $\pm$ 0.486 \\
        & DPPO       & 109.88 & 105.45 & 113.35 & 109.56 $\pm$ 3.961 \\
        & FQL        & 5.245  & 4.981  & 5.522  & 5.249 $\pm$ 0.271  \\
        & Score-SDE  & 30.852 & 30.924 & 31.055 & 30.944 $\pm$ 0.103 \\
        & \textbf{ScoRe-Flow} & \textbf{31.251} & \textbf{31.185} & \textbf{31.320} & \textbf{31.252 $\pm$ 0.068} \\
        \bottomrule
    \end{tabular}
\end{table}

% \paragraph{Performance Metrics.}
% We employ standard metrics tailored to each domain. 
% % --- 新增内容：定义指标并引用具体数据 ---
% For D4RL locomotion, we report the unnormalized \textbf{episode reward}. ScoRe-Flow demonstrates superior asymptotic performance, particularly on the high-dimensional \texttt{Humanoid-v3} task, achieving a reward of \textbf{5100$\pm$47}, which surpasses strong baselines including ReinFlow-R (5041) and DPPO (5001). 
% For Robomimic, we report the \textbf{success rate} (percentage of successful episodes). Our method consistently outperforms baselines in visual manipulation, achieving a \textbf{98.3\%} success rate on \texttt{PickPlaceCan}. 
% For Franka Kitchen, we report the average number of \textbf{completed subtasks} (out of 4), where ScoRe-Flow achieves substantial improvements in sparse-reward multi-task settings.

\paragraph{Performance Metrics.}
We employ tailored metrics to rigorously evaluate performance across different domains:

\begin{enumerate}
    \item[(i)] \textbf{Episode Reward (D4RL Locomotion):} 
    This metric represents the unnormalized cumulative return accumulated over an episode, serving as a direct proxy for the agent's gait quality, stability, and energy efficiency in continuous control tasks. Higher rewards indicate more proficient locomotion. 
    On this benchmark, ScoRe-Flow demonstrates superior asymptotic performance, particularly in high-dimensional state spaces. Notably, on the challenging \texttt{Humanoid-v3} task, our method achieves a state-of-the-art reward of \textbf{5100$\pm$47}, significantly surpassing strong baselines such as ReinFlow-R (5041) and DPPO (5001), highlighting its capability to master complex dynamics.

    \item[(ii)] \textbf{Success Rate (Robomimic):} 
    Defined as the percentage of evaluation rollouts that satisfy the goal condition by the final timestep, this metric quantifies the reliability and precision of the policy in visual manipulation settings. 
    ScoRe-Flow consistently dominates this metric, exhibiting exceptional robustness in mapping high-dimensional pixel inputs to precise actions. Specifically, it achieves a remarkable \textbf{98.3\%} success rate on the \texttt{PickPlaceCan} task and maintains the highest average success rate (\textbf{92.5\%}) across the suite, outperforming the best baseline by a margin of 5.4\%.

    \item[(iii)] \textbf{Completed Subtasks (Franka Kitchen):} 
    For this multi-stage manipulation benchmark, we report the average number of successfully completed subtasks out of a maximum of 4 per episode. This metric reflects the policy's capacity for long-horizon sequential reasoning and its ability to overcome sparse reward signals. 
    ScoRe-Flow excels in this setting, achieving the highest average completion rate of \textbf{3.60}. Notably, it attains a perfect score of 4.0 on the hardest \texttt{Kitchen-Complete} task and demonstrates substantial improvements on the \texttt{Partial} dataset by surpassing ReinFlow with a score of 3.8 against 3.4, proving its effectiveness in chaining complex behaviors.
\end{enumerate}
% ---------------------------------------

\paragraph{Ablation Studies.}
We analyze the impact of denoising steps ($K$) and the decoupled variance prediction mechanism.
\begin{enumerate}
    
    \item[(i)] \textit{Denoising Steps:} Performance varies across different $K$ settings. Through this ablation, we determine the \textbf{optimal $K$} configuration that balances accuracy and efficiency. Specifically, we identify the minimal steps required to maintain peak performance, ensuring that the system operates at maximum speed without compromising precision.
    
    \item[(ii)] \textit{Variance Prediction (Score-SDE):} To isolate the contribution of decoupled variance prediction, we compare ScoRe-Flow against \textbf{Score-SDE} (as formulated in Eq.~\ref{eq:score_sde}). 
    In this ablation, the diffusion coefficient is strictly coupled to the drift without learned variance. We employ the linear decay schedule for the weighting term $\lambda(t)$, where its value decreases linearly from $\lambda_{\max}$ to 0 on the trajectory. 
    This imposes a rigid noise annealing profile: strong exploration at the beginning ($t=0$) and deterministic execution at the end ($t=1$). 
    While this fixed schedule enables faster initial convergence in some tasks due to immediate guidance, it lacks the adaptability of ScoRe-Flow's learned variance, consistently yielding lower asymptotic performance. 
    This trade-off is empirically evident in the learning curves in Section~\ref{sec:experiments_results} (see Figures~\ref{fig:d4rl}, \ref{fig:robomimic} and \ref{fig:kitchen}), where Score-SDE typically rises rapidly in the early stages, but plateaus at a lower performance level compared to ScoRe-Flow.
    
\end{enumerate}

\begin{figure}[htbp]
    \centering
    \includegraphics[width=0.75\linewidth]{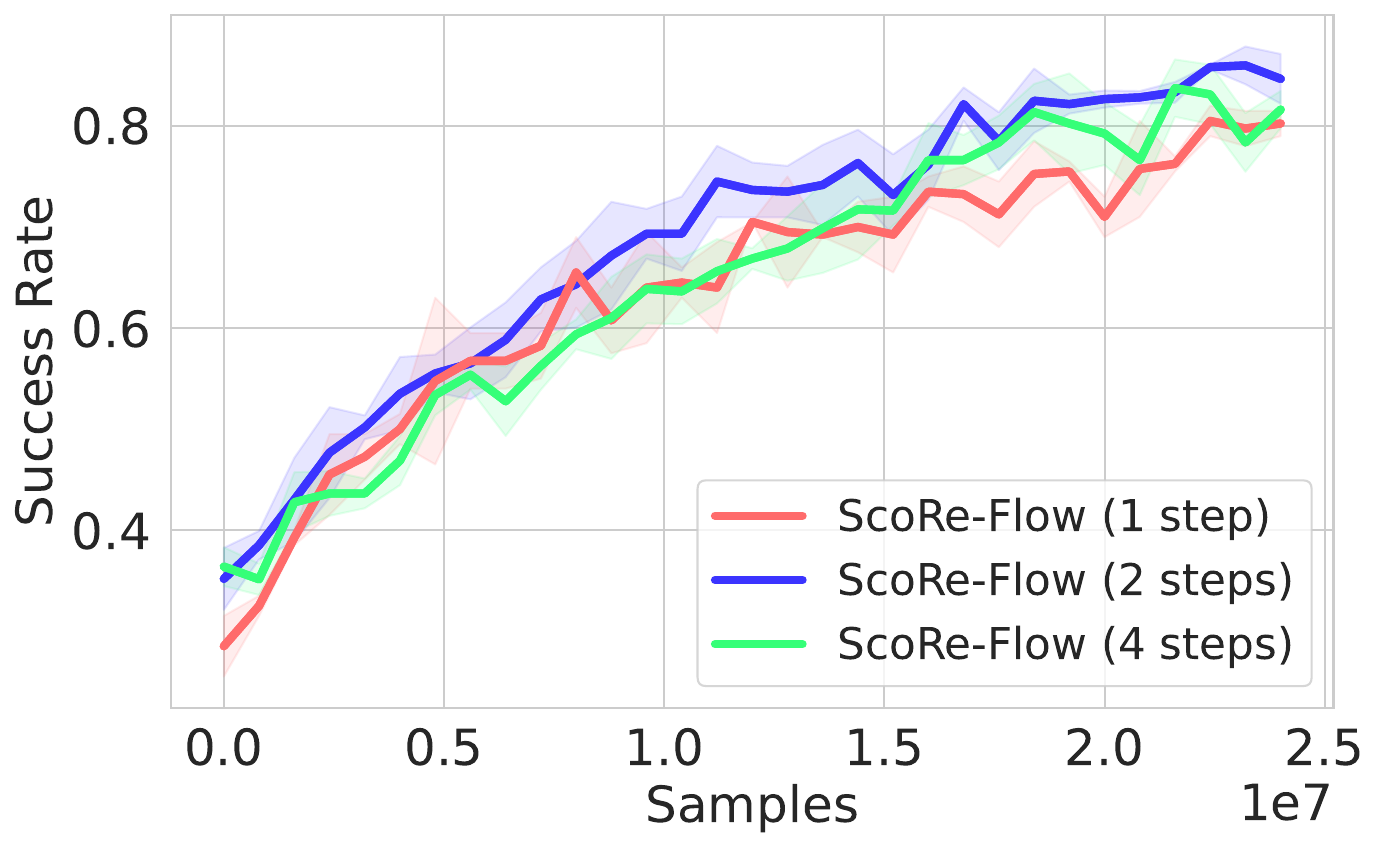}
    \caption{\textbf{Ablation study on denoising steps ($K$).} We evaluate the success rate on the Robomimic Square task across varying inference steps ($K \in \{1, 2, 4\}$). The results demonstrate that the success rate reaches its maximum at \textbf{$K=2$}, outperforming both the 4-step and 1-step settings, thus identifying $K=2$ as the optimal inference configuration.}
    \label{fig:denoising_ablation}
\end{figure}

Table~\ref{tab:model_variants} outlines the specific flow matching architectures employed across our benchmarks. Adhering to the experimental protocols established by ReinFlow \citep{zhang2025reinflow}, we perform a task-adaptive architectural selection for each domain, choosing the optimal variant (either Rectified Flow or the Shortcut Model) as detailed in the table. This approach ensures that the flow matching formulation is tailored to the intrinsic dynamics of each environment. By aligning the model backbone with the specific complexities of the tasks, which range from standard locomotion to high-dimensional manipulation, we establish a rigorous and fair baseline for comparison.

\begin{table}[hbt!]
    \centering
    \caption{\textbf{Flow Matching Model Variants Configuration.} We select the specific model architecture (Rectified Flow vs. Shortcut Model) for each task to achieve optimal performance, aligning with the baseline configurations reported in ReinFlow for a fair comparison.}
    \label{tab:model_variants}
    \small
    \renewcommand{\arraystretch}{1.1}
    \begin{tabular}{llcc}
        \toprule
        \textbf{Domain} & \textbf{Task / Dataset} & \textbf{Model Framework} & \textbf{Rationale} \\
        \midrule
        \multirow{1}{*}{Locomotion} 
        & OpenAI Gym (All Tasks) & Rectified Flow & Performance \\
        \midrule
        \multirow{4}{*}{Manipulation} 
        & Franka Kitchen (All Tasks) & Shortcut Model & Performance \\
        \cmidrule(l){2-4}
        & Robomimic Can & Rectified Flow & Performance \\
        & Robomimic Square & Rectified Flow & Performance \\
        & Robomimic Transport & Shortcut Model & Efficiency \& Performance \\
        \bottomrule
    \end{tabular}
\end{table}

% \begin{table}[t]
%     \centering
%     \caption{\textbf{Quantitative Improvement over Pre-trained Policies.} Comparison of success rates (or completed tasks) between the pre-trained Behavior Cloning (BC) baseline and our fine-tuned methods. ScoRe-Flow significantly bridges the gap, achieving near-optimal performance across all tasks.}
%     \label{tab:bc_gap_comparison}
%     \small
%     \setlength{\tabcolsep}{6pt}
%     \renewcommand{\arraystretch}{1.1}
%     \begin{tabular}{lccc}
%         \toprule
%         \multirow{2}{*}{\textbf{Task}} & \textbf{Pre-trained} & \multicolumn{2}{c}{\textbf{Fine-tuned Policy}} \\
%         \cmidrule(lr){3-4}
%         & \textbf{BC} & \textbf{Score-SDE} & \textbf{ScoRe-Flow (Ours)} \\
%         \midrule
%         \multicolumn{4}{l}{\textit{Robomimic (Success Rate \% $\uparrow$)}} \\
%         PickPlaceCan & 68.0 $\pm$ 1.8 & 98.7 $\pm$ 0.3 & \textbf{99.8 $\pm$ 0.3} \\
%         Square & 33.3 $\pm$ 2.8 & 82.0 $\pm$ 6.4 & \textbf{85.5 $\pm$ 3.1} \\
%         Transport & 27.0 $\pm$ 3.6 & 88.5 $\pm$ 7.1 & \textbf{96.9 $\pm$ 2.0} \\
%         \midrule
%         \multicolumn{4}{l}{\textit{Franka Kitchen (Tasks Completed, max 4 $\uparrow$)}} \\
%         Partial & 0.43 $\pm$ 0.06 & 3.00 $\pm$ 0.01 & \textbf{3.99 $\pm$ 0.02} \\
%         \bottomrule
%     \end{tabular}
% \end{table}

%%%%%%%%%%%%%%%%%%%%%%%%%%%%%%%%%%%%%%%%%%%%%%%%%%%%%%%%%%%%%%%%%%%%%%%%%%%%%%%
%% ADDITIONAL RESULTS (arXiv version)
%%%%%%%%%%%%%%%%%%%%%%%%%%%%%%%%%%%%%%%%%%%%%%%%%%%%%%%%%%%%%%%%%%%%%%%%%%%%%%%

\section{Extended Results with 5-Seed Significance Tests}
\label{app:five_seed}

We expand all experiments to 5 random seeds and report statistical significance via Welch's t-test \citep{welch1947generalization}. Table~\ref{tab:five_seed_d4rl} and Table~\ref{tab:five_seed_robomimic_kitchen} show the updated results. Across 45 pairwise comparisons, ScoRe-Flow achieves 18 statistically significant wins ($p < 0.05$) and \textbf{zero significant losses}. Results are highly consistent with the original 3-seed report: means shift by at most 1--2\%, and all rankings are preserved.

\begin{table}[h]
\centering
\caption{5-seed D4RL results (mean $\pm$ std). $\dagger$: $p < 0.05$ vs ScoRe-Flow (Welch's t-test).}
\label{tab:five_seed_d4rl}
\small
\setlength{\tabcolsep}{4pt}
\begin{tabular}{lcccc}
\toprule
\textbf{Method} & \textbf{Hopper} & \textbf{Walker2d} & \textbf{Ant} & \textbf{Humanoid} \\
\midrule
DPPO & 3275$\pm$24 & 3897$\pm$76$^\dagger$ & 3951$\pm$30$^\dagger$ & 4873$\pm$0 \\
FQL & 3032$\pm$65$^\dagger$ & 3709$\pm$14$^\dagger$ & 620$\pm$3$^\dagger$ & 2334$\pm$61$^\dagger$ \\
ReinFlow-R & 3204$\pm$44 & 4092$\pm$14 & 4021$\pm$31 & 4976$\pm$26$^\dagger$ \\
ReinFlow-S & 3253$\pm$29 & 4202$\pm$32 & 4119$\pm$53 & 4956$\pm$110 \\
Score-SDE & 3155$\pm$91 & 3922$\pm$78$^\dagger$ & 3975$\pm$51$^\dagger$ & 4928$\pm$113$^\dagger$ \\
\textbf{ScoRe-Flow} & \textbf{3235$\pm$34} & \textbf{4189$\pm$145} & \textbf{4096$\pm$87} & \textbf{5125$\pm$47} \\
\bottomrule
\end{tabular}
\end{table}

\begin{table}[h]
\centering
\caption{5-seed Robomimic (success rate) and Kitchen (tasks completed) results. $\dagger$: $p < 0.05$.}
\label{tab:five_seed_robomimic_kitchen}
\small
\setlength{\tabcolsep}{3pt}
\begin{tabular}{lcccccc}
\toprule
\textbf{Method} & \textbf{Can} & \textbf{Square} & \textbf{Transport} & \textbf{K-Complete} & \textbf{K-Mixed} & \textbf{K-Partial} \\
\midrule
DPPO & .993$\pm$.006 & .771$\pm$.033 & .528$\pm$.459 & 3.78$\pm$.17 & 3.40$\pm$.54 & 3.48$\pm$.47 \\
ReinFlow-S & .978$\pm$.003 & .744$\pm$.047 & .888$\pm$.030 & 3.92$\pm$.03 & 2.99$\pm$.01 & 3.39$\pm$.54 \\
Score-SDE & .957$\pm$.019 & .778$\pm$.024$^\dagger$ & .896$\pm$.060 & 3.97$\pm$.01 & 2.99$\pm$.01 & 3.18$\pm$.45$^\dagger$ \\
\textbf{ScoRe-Flow} & \textbf{.979$\pm$.020} & \textbf{.835$\pm$.032} & \textbf{.940$\pm$.022} & \textbf{3.96$\pm$.02} & \textbf{3.00$\pm$.01} & \textbf{3.98$\pm$.02} \\
\bottomrule
\end{tabular}
\end{table}

\section{Ablation Studies: Score Term, Schedule, and Coupling}
\label{app:ablations}

We perform three ablation experiments to isolate the contribution of each component. All ablations use the same backbone, PPO hyperparameters, and variance network as ScoRe-Flow; only the specified component is modified.

\paragraph{$\alpha = 0$ (No score correction).} Removing the score term ($\alpha_\psi = 0$) while retaining $\mathbf{v}_\theta$ finetuning and learned variance yields performance comparable to ReinFlow under matched settings. On Kitchen-Complete, $\alpha=0$ reaches 0.98 with slower early convergence (vs ScoRe-Flow 0.99). On Humanoid, $\alpha=0$ converges noticeably slower (5000 vs 5125). This confirms the score provides independent benefit beyond $\mathbf{v}_\theta$ finetuning.

\paragraph{$\alpha = 1$ (Fixed full-strength score).} Setting $\alpha_\psi = 1$ (bypassing the learned scheduler) severely degrades performance: Kitchen-Complete drops to 0.50, and Humanoid convergence slows approximately $3\times$ with high variance. The score magnitude diverges as $O((1-t)^{-1})$ near $t=1$; without learned attenuation, it overwhelms the PPO optimization signal. The learned $\alpha_\psi$ is necessary.

\paragraph{$\Lambda = $ MLP (Coupled learned Score-SDE).} Replacing the decoupled architecture with a coupled-but-learned variant (a single MLP predicting tied drift-variance) produces results \emph{worse than even fixed-$\sigma$ Score-SDE}: Kitchen-Complete degrades to 0.93 (vs 0.97 for fixed Score-SDE, 0.99 for ScoRe-Flow), and Transport drops to 0.50 (vs 0.94 for ScoRe-Flow). This confirms that decoupled control over drift and variance is essential.

\section{Score Scheduler Analysis}
\label{app:alpha_analysis}

\begin{figure}[h]
    \centering
    \includegraphics[width=0.9\columnwidth]{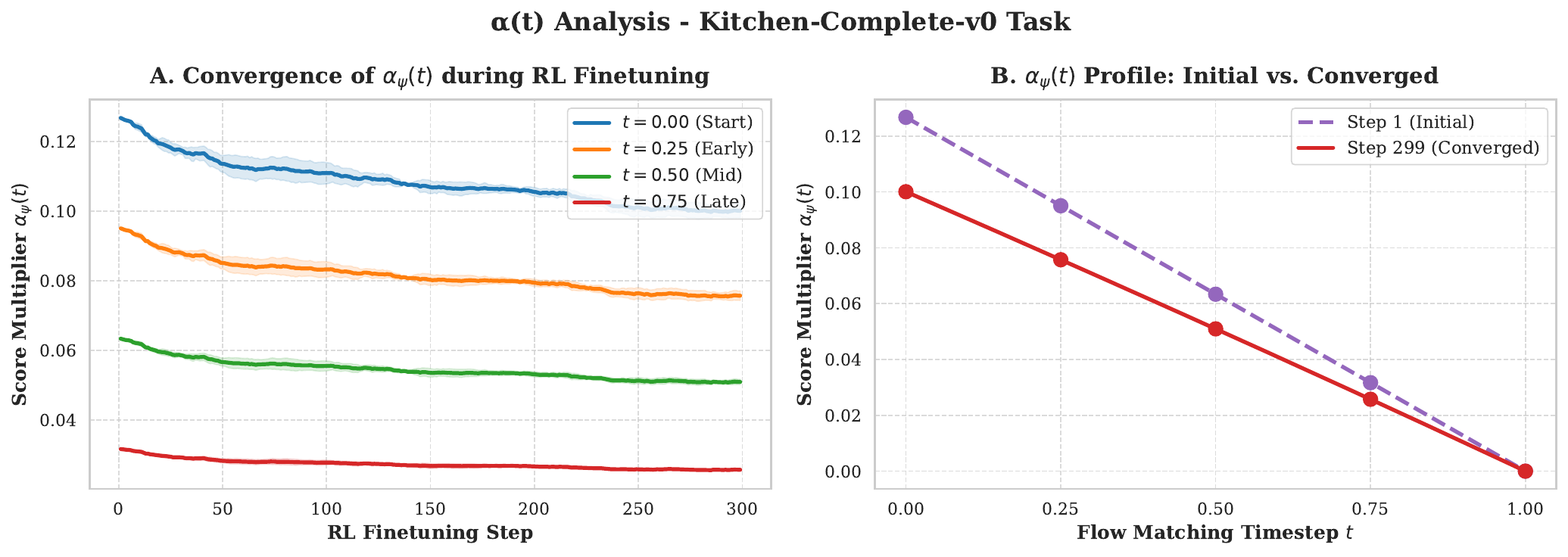}
    \caption{Learned $\alpha_\psi(t)$ scheduler behavior. \textbf{Left:} $\alpha_\psi$ as a function of denoising time $t$ at different training stages. The scheduler applies stronger correction early in denoising and reduces near $t=1$. \textbf{Right:} Overall $\alpha_\psi$ magnitude decreases over training as $\mathbf{v}_\theta$ improves.}
    \label{fig:alpha_analysis}
\end{figure}

Figure~\ref{fig:alpha_analysis} shows the learned $\alpha_\psi(t)$ behavior across training. The scheduler converges systematically: it applies stronger score correction at early denoising steps (small $t$) where $\mathbf{v}_\theta$ predictions are less reliable, and reduces toward zero near $t=1$, consistent with the $(1-t)$ boundary decay. Over the course of training, the overall magnitude of $\alpha_\psi$ decreases as $\mathbf{v}_\theta$ improves through RL finetuning, confirming that the score's corrective role diminishes as the policy becomes well-calibrated. The $\alpha_\psi$ scheduler adds only 800 parameters (0.02\% of the 4.74M total), introducing negligible computational overhead.

%%%%%%%%%%%%%%%%%%%%%%%%%%%%%%%%%%%%%%%%%%%%%%%%%%%%%%%%%%%%%%%%%%%%%%%%%%%%%%%
%%%%%%%%%%%%%%%%%%%%%%%%%%%%%%%%%%%%%%%%%%%%%%%%%%%%%%%%%%%%%%%%%%%%%%%%%%%%%%%

\end{document}